\documentclass[10pt,twocolumn,letterpaper]{article}

 \usepackage[preprint]{cvpr}      

\usepackage{todonotes}
\usepackage{amsthm}
\usepackage{mathtools}
\usepackage{algorithm}
\usepackage{algpseudocode}
\algrenewcommand\alglinenumber[1]{\footnotesize #1}
\usepackage{multirow}
\usepackage[table]{xcolor}
\definecolor{trailmixgreen}{RGB}{207,237,192}

\usepackage{aliascnt}

\newaliascnt{proposition}{theorem}

\aliascntresetthe{proposition}
\newaliascnt{lemma}{theorem}
\newtheorem{lemma}[lemma]{Lemma}
\aliascntresetthe{lemma}
\newaliascnt{corollary}{theorem}

\aliascntresetthe{corollary}

\AtBeginDocument{%
  \crefname{theorem}{Theorem}{Theorems}%
  \crefname{proposition}{Proposition}{Propositions}%
  \crefname{lemma}{Lemma}{Lemmas}%
  \crefname{corollary}{Corollary}{Corollaries}%
  \crefname{assumption}{Assumption}{Assumptions}%
  \crefname{algorithm}{Algorithm}{Algorithms}%
  \crefname{equation}{Eq.}{Eqs.}%
}

\newcommand{\KL}{\mathrm{KL}}
\newcommand{\E}{\mathbb{E}}

\newcommand{\dsGen}{\textsc{GenEval}}
\newcommand{\dsEdit}{\textsc{SEAM-Bench}}

\makeatletter
\newbox\tt@parkedlabels
\newif\iftt@parkedinlabel
\pretocmd{\@outputpage}{%
  \global\setbox\tt@parkedlabels\box\@labels
  \if@inlabel\global\tt@parkedinlabeltrue\else\global\tt@parkedinlabelfalse\fi
  \global\@inlabelfalse}{}{\PackageWarning{preamble}{Could not patch \string\@outputpage}}
\AddToHook{shipout/after}{%
  \global\setbox\@labels\box\tt@parkedlabels
  \iftt@parkedinlabel\global\@inlabeltrue\fi}
\makeatother

\definecolor{cvprblue}{rgb}{0.21,0.49,0.74}
\usepackage[pagebackref,breaklinks,colorlinks,allcolors=cvprblue]{hyperref}

\def\confName{CVPR}
\def\confYear{2027}

\newcommand{\methodname}{\textsc{TrailMix}}

\title{Diffusion Steering via Adaptive Reward Estimation and Resampling}

\author{
  Abinitha Gourabathina\textsuperscript{1} \quad
  Sana Tonekaboni\textsuperscript{1,2} \quad
  Marzyeh Ghassemi\textsuperscript{1} \quad
  Akbar Rafiey\textsuperscript{2,3}\\[2pt]
  \textsuperscript{1}Massachusetts Institute of Technology \quad
  \textsuperscript{2}RBC Borealis \quad
  \textsuperscript{3}NYU\\
  {\tt\small \{abinitha, mghassem\}@mit.edu} \quad
  {\tt\small sana.tonekaboni@rbc.com} \quad
  {\tt\small akbar.rafiey@rbc.com}
}

\begin{document}
\maketitle
\begin{abstract}
Text-to-image diffusion models can generate images that satisfy complex prompts, yet often produce images that omit requested objects, attributes, or relationships. We ask if models can more reliably generate text-aligned images by redistributing probability toward their more successful outcomes. We propose \methodname{}, a training-free method that uses the base model's distribution of possible outcomes to adaptively guide generation toward better text alignment. During denoising, \methodname{} generates inexpensive lookahead images from parallel trajectories (trails) and scores their alignment with the prompt using an external judge. The resulting distribution of reward scores allows us to estimate how strongly to steer generation toward a desired alignment level. We use these estimates to  reweight and resample trajectories, favoring more promising outcomes. Across three text-to-image benchmarks, \methodname{} outperforms existing baselines, with particularly strong gains on rare multi-object compositions. We further extend \methodname{} to targeted image editing and introduce \textsc{SEAM-Bench} (Semantic Editing and Attribute Maintenance), a systematic benchmark evaluating whether semantic edits preserve other image content. Experiments on SEAM-Bench and existing benchmark PIE-Bench demonstrate the effectiveness of our approach for controlled image editing. Our results show how a model's own possible outcomes can inform  reliable and controllable generation.
\end{abstract}
\section{Introduction}
\label{sec:intro}

\begin{figure*}[!h]
    \centering
    \includegraphics[width=\textwidth]{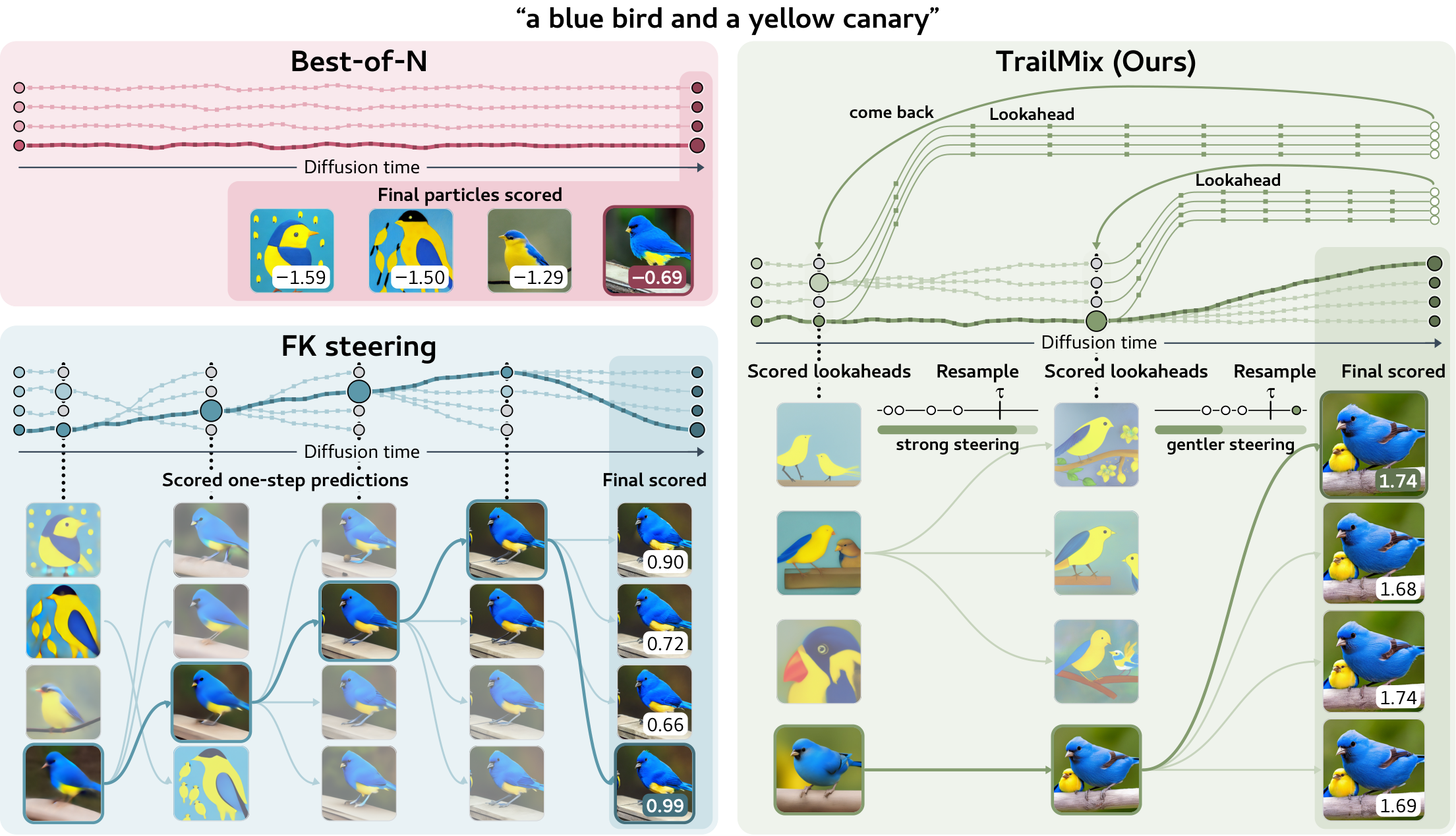}
    \caption{\textbf{Method overview.}
For the prompt ``a blue bird and a yellow canary,'' we compare three strategies. Best-of-$N$ selects the highest-scoring final sample, while Feynman-Kac (FK) resamples particles during denoising using one-step reward predictions. \methodname{} uses inexpensive lookahead previews to identify promising trajectories and adaptive steering strength to produce the most aligned images.}
    \label{fig:teaser}
\end{figure*}

Text-to-image diffusion models have shown impressive performance across tasks~\cite{rombach2022ldm,podell2024sdxl}, yet they can still fail to produce images that satisfy specific semantic requirements of a query ~\cite{ghosh2023geneval,huang2025t2icompbench,park2025r2f, zarei2024improving}, such as counts or compositions of multiple objects.  

Reward guidance has shown promise in improving image-text alignment via a reward function that scores how well an image satisfies its prompt. This function is often implemented using a pretrained reward model~\cite{xu2023imagereward} or a vision-language model judge~\cite{sargent2026vlic}. Reward-based finetuning methods optimize diffusion models toward preferred outputs ~\cite{black2024ddpo,fan2023dpok,clark2024draft,wallace2024diffusiondpo}, but require model optimization which can degrade general capabilities \cite{dong2024continually}. Inference-time approaches instead leverage reward information without modifying the pretrained model. Best-of-$N$ sampling generates independent candidates and selects the highest-reward image. However, it  allocates a large amount of computation to completing low-reward, misaligned candidates that are ultimately discarded~\cite{chatterjee2018sample}.

Recent methods use reward information during the generation process itself. Particle-based sampling and search methods, like Feynman-Kac (FK) steering, evaluate intermediate trajectories and allocate computation toward promising outcomes~\cite{li2024svdd,singhal2025fk,zhang2026classical,guimaraes2026psp}. However, noisy intermediate previews may not yet reveal the semantic details needed to assess alignment with complex prompts~\cite{tinaz2026emergenceevolutioninterpretableconcepts,yue2024exploringdiffusiontimestepsunsupervised,guo2026earlyqualityassessmenttexttoimage}. Guidance-based methods take a complementary approach, incorporating reward-derived gradients into denoising updates~\cite{bansal2023universalguidancediffusionmodels}. For example, LiDAR~\cite{kim2026lookahead} uses a few coarse denoising steps to generate inexpensive \emph{lookahead images}, whose rewards inform an analytic correction to the denoising score.
However, we identify another benefit of lookahead images. By revealing semantic details that are difficult to assess in noisy intermediate previews, these images allow us to better estimate the alignment of potential outcomes when resampling trajectories.

We therefore propose \methodname{}, a training-free method that adaptively combines the pretrained model's generative distribution with reward-based selection to improve image-text alignment (Figure~\ref{fig:teaser}). Building on FK steering~\cite{singhal2025fk}, we run multiple denoising trajectories (\emph{trails}) in parallel and periodically resample them using rewards computed from lookahead images. Beyond improving expected reward estimation, we adapt how strongly resampling favors higher-reward trails. 
The model's own possible generations thus inform both trajectory selection and how strongly rewards influence sampling. 

We evaluate \methodname{} on \textsc{GenEval}~\cite{ghosh2023geneval}, \textsc{T2I-CompBench++}~\cite{huang2025t2icompbench}, and \textsc{RareBench}~\cite{park2025r2f}, using Stable Diffusion v1.4 (SDv1.4) and Stable Diffusion XL (SDXL). \methodname{} consistently improves prompt alignment over four-particle baselines across both models, with the largest gains on rare multi-object compositions, where it outperforms Best-of-8 using half as many particles.

The same steering mechanism also extends to targeted image editing. We introduce \dsEdit{} (Semantic Editing and Attribute Maintenance Benchmark), a systematic, attribute-level benchmark that jointly evaluates edit success and preservation of unrelated image details, including edits toward and against stereotypical associations. \methodname{} achieves the highest edit success among evaluated methods on \dsEdit{} across both diffusion models, while maintaining strong source preservation. Its benefits also extend to existing images on PIE-Bench~\cite{ju2024pnp}, particularly with SDXL, and to counter-stereotypical editing, where it achieves more balanced success across editing directions.

\section{Related Work}
\label{sec:related}

\noindent
\textbf{Finetuning-based alignment.}
Finetuning approaches update diffusion model parameters to improve alignment with reward objectives. DDPO and DPOK~\cite{black2024ddpo,fan2023dpok} optimize rewards through reinforcement learning, DRaFT~\cite{clark2024draft} backpropagates reward gradients through sampling, and Diffusion-DPO~\cite{wallace2024diffusiondpo} learns from preference pairs. However, over-optimizing rewards can increase scores at the expense of image quality and diversity~\cite{lamba2025alignment,jena2025elucidating}. We focus on inference-time steering, which adapts generation without further model training. These approaches broadly include guidance-based methods and sampling methods.

\vspace{2pt}
\noindent
\textbf{Guidance-based methods.}
SEGA~\cite{brack2023segainstructingtexttoimagemodels} steers generation along text-defined semantic directions, while Prompt-to-Prompt~\cite{hertz2022prompt} controls cross-attention maps to edit image content while preserving its layout. Universal Guidance~\cite{bansal2023universalguidancediffusionmodels} and FreeDoM~\cite{yu2023freedomtrainingfreeenergyguidedconditional} adjust denoising updates using gradients of differentiable objectives. LiDAR~\cite{kim2026lookahead} uses inexpensive lookahead images and their rewards to compute an analytic guidance correction.

\noindent
\textbf{Sampling and search methods.}
Sampling and search methods allocate computation across candidate trajectories. Best-of-$N$ selects among completed images, while noise search refines noise inputs, PSP prunes candidates, and DTS reuses completed rollouts to guide tree exploration~\cite{ma2025inferencescaling,guimaraes2026psp,jain2025dts}. TDS and DAS apply sequential Monte Carlo to conditional generation and reward alignment, respectively~\cite{wu2023tds,kim2025das}, while SVDD selects transitions using estimated future rewards~\cite{li2024svdd}. Feynman-Kac (FK) steering~\cite{singhal2025fk} maintains parallel denoising trajectories and resamples them at intermediate steps using reward-based weights. Building on this framework, \methodname{} evaluates trajectories through short lookahead continuations and uses the resulting reward distribution to adapt resampling.
\section{Method}
\label{sec:method}

\methodname{} steers a pretrained diffusion model toward images that better satisfy the semantic requirements of a text prompt. It runs multiple denoising trajectories in parallel and uses inexpensive lookahead previews to assess their possible outcomes before generation is complete (see \Cref{fig:teaser}). A reward judge scores these previews for alignment with the prompt, and the resulting scores determine both how strongly to steer and which trajectories to favor (see \Cref{alg:tailtilt}). We first define the reward-guided target distribution (\cref{sec:target}) and introduce lookahead previews (\cref{sec:lookahead}), then describe adaptive steering strength (\cref{sec:strength}) and its implementation through reward estimation and resampling (\cref{sec:twist}).

\subsection{Target distribution}
\label{sec:target}

For a prompt $c$, let $p_\theta(x_0\mid c)$ denote the pretrained model's distribution over completed images $x_0$. An external judge assigns each image a reward $r(x_0,c)$ measuring how well it satisfies the prompt. We define the reward-guided target distribution by
\begin{equation}
\label{eq:target}
p_{\beta_c}(x_0\mid c)
\propto p_\theta(x_0\mid c)e^{\beta_c r(x_0,c)}.
\end{equation}
This reweights the base distribution toward higher-reward images, with the \emph{steering strength} $\beta_c\ge0$ controlling the shift. Setting $\beta_c=0$ recovers the base model, while larger strengths increasingly favor high-reward outcomes.

\subsection{Lookahead previews}
\label{sec:lookahead}

We maintain $n$ denoising trajectories, or \emph{particles}, initialized independently as $x_T^i\sim p(x_T)$ for $i=1,\ldots,n$. The particles follow the pretrained denoising process between selected checkpoints $t_1>\cdots>t_K$. At checkpoint $t_k$, let $\{x_{t_k}^i\}_{i=1}^n$ denote their current states. Prior diffusion steering methods estimate rewards from intermediate predictions of the final image~\cite{singhal2025fk}, but these predictions may lack the semantic details needed to assess prompt alignment.

We instead generate a lookahead preview $\tilde x_{0,k}^i$ from each particle using $L$ coarse denoising steps, leaving the original states unchanged. We evaluate each preview using the same reward function that defines the target distribution:
\begin{equation}
\label{eq:preview}
s_k^i=r(\tilde x_{0,k}^i,c).
\end{equation}
The previews provide inexpensive estimates of potential final rewards, which we use to adapt the steering strength and guide resampling.

\begin{algorithm}[!h]
\caption{\methodname{}}
\label{alg:tailtilt}
\begin{algorithmic}[1]
\Require prompt $c$, model $p_\theta$, reward $r$, target mean reward $\tau$, $n$ particles, preview length $L$, checkpoints $(t_1,\ldots,t_K)$
\State Draw $x_T^i\sim p(x_T)$ independently and set $h^i\gets0$ for $i=1,\ldots,n$
\For{$k=1,\ldots,K$}
  \State Denoise each particle to $t_k$ using the base model
  \State Generate an $L$-step preview $\tilde x_{0,k}^i$ from each $x_{t_k}^i$
  \State Set $s_k^i\gets r(\tilde x_{0,k}^i,c)$
  \State Compute $\hat\beta_{c,k}$ from $\{(s_k^i,h^i)\}_{i=1}^n$ (\cref{eq:level-estimate})
  \State Resample $n$ particles with probabilities $\omega_k^i(\hat\beta_{c,k})$ (\cref{eq:weights})
  \State Each selected copy of parent $i$ inherits $h\gets\hat\beta_{c,k}s_k^i$
\EndFor
\State Complete denoising, evaluate final rewards, and return the highest-reward image
\end{algorithmic}
\end{algorithm}

\subsection{Adaptive steering strength}
\label{sec:strength}

We adapt the steering strength to account for how readily the pretrained model produces well-aligned images. Prompts where such images are already likely require less reweighting, while others may need stronger steering. We therefore choose the smallest strength that achieves a target mean reward $\tau$ under the reward-guided distribution. For any candidate strength $\beta\ge0$, let $p_\beta$ denote \cref{eq:target} with $\beta_c$ replaced by $\beta$, and define
\begin{equation}
\label{eq:mean-reward}
\mu_c(\beta)=\E_{x_0\sim p_\beta(\cdot\mid c)}[r(x_0,c)].
\end{equation}
As $\beta$ increases, higher-reward images receive greater relative probability, so $\mu_c(\beta)$ is nondecreasing. We select
\begin{equation}
\label{eq:level}
\beta_c=\inf\{\beta\ge0:\mu_c(\beta)\ge\tau\}.
\end{equation}
This choice also has a minimum-change interpretation: when the target is attainable at a finite strength, $p_{\beta_c}$ is the distribution closest to the base model in KL divergence among those satisfying the reward constraint (App. \ref{sec:supp-theory}):
\begin{equation}
\label{eq:minimal-change}
p_{\beta_c}(\cdot\mid c)
=\operatorname*{argmin}_{Q:\,\E_{x_0\sim Q}[r(x_0,c)]\ge\tau}
\KL\!\left(Q\,\|\,p_\theta(\cdot\mid c)\right).
\end{equation}

\subsection{Reward estimation and resampling}
\label{sec:twist}

The preceding formulation defines the desired steering strength using the distribution of completed images. In practice, \methodname{} estimates this strength from lookahead rewards at each checkpoint and uses it to resample particles. Since earlier resampling may have favored certain trajectories, we update their selection weights to reflect the change in reward preference. Each particle retains the reward-weighted score $h_{k-1}^i$ inherited from its parent at the previous checkpoint, with $h_0^i=0$. For a candidate strength $\beta$, the normalized weights are
\begin{equation}
\label{eq:weights}
\omega_k^i(\beta)
\propto \exp\!\left(\beta s_k^i-h_{k-1}^i\right),
\qquad \sum_{i=1}^n\omega_k^i(\beta)=1.
\end{equation}
The difference compares the current reward-weighted score with the score already applied at the previous checkpoint, so resampling can incorporate new reward estimates without repeatedly amplifying earlier selection.

We use these weights to estimate the mean reward for each candidate strength and select the smallest value that reaches the target $\tau$:
\begin{equation}
\label{eq:level-estimate}
\hat\beta_{c,k}
=\inf\left\{\beta\ge0:
\sum_{i=1}^n\omega_k^i(\beta)s_k^i\ge\tau
\right\}.
\end{equation}
We then resample $n$ particles with replacement according to $\omega_k^i(\hat\beta_{c,k})$, so some trajectories are copied while others are discarded. Each copy of a selected particle $i$ inherits its checkpoint state and the updated score $h=\hat\beta_{c,k}s_k^i$, then continues denoising independently. After the final checkpoint, we complete all trajectories, evaluate their final rewards, and return the highest-reward image.

\section{Experiments}
\label{sec:exp}

We evaluate \methodname{} on text-to-image generation and image editing. For both tasks, we use two pretrained diffusion models: Stable Diffusion v1.4 (SDv1.4)~\cite{rombach2022ldm} and Stable Diffusion XL (SDXL)~\cite{podell2024sdxl}. We provide all implementation details and code to reproduce results in App. \ref{sec:supp-impl}.

\subsection{Text-to-image generation}
\label{sec:t2i}

\begin{table*}[!h]
\centering
\footnotesize
\setlength{\tabcolsep}{3pt}
\renewcommand{\arraystretch}{1.12}
\resizebox{0.82\textwidth}{!}{%
\begin{tabular}{@{}ll c cc cccc ccc c@{}}
\toprule
& & & \multicolumn{2}{c}{\textsc{RareBench}} & \multicolumn{4}{c}{\dsGen} & \multicolumn{3}{c}{\textsc{T2I-CompBench++}} & \\
\cmidrule(lr){4-5}\cmidrule(lr){6-9}\cmidrule(lr){10-12}
Model & Method & \shortstack{Particles\\$n$} & \shortstack{Single\\obj.} & \shortstack{Multi\\obj.} & Overall & Counting & Position & \shortstack{Color\\attr.} & Binding & Spatial & Numeracy & \shortstack{U-Net /\\judge calls}\\
\midrule
\multirow{14}{*}{\rotatebox[origin=c]{90}{SD v1.4}} & Base model (single) & 1 & 51.5 & 37.0 & 45.3 & 35.0 & 5.0 & 9.0 & 39.6 & 11.5 & 46.0 & 200\,/\,1\\
\cmidrule(lr){2-13}
 & Universal Guidance~\cite{bansal2023universalguidancediffusionmodels} & - & 54.4 & 43.3 & 40.4 & 36.2 & 6.0 & 6.0 & 39.5 & 9.1 & 43.4 & 400\,/\,100\\
 & FreeDoM~\cite{yu2023freedomtrainingfreeenergyguidedconditional} & - & 64.6 & 49.4 & 38.2 & 25.8 & 4.0 & 2.0 & 42.2 & 1.6 & 33.8 & 520\,/\,100\\
 & SEGA~\cite{brack2023segainstructingtexttoimagemodels} & - & 54.0 & 41.6 & 43.5 & 35.0 & 3.0 & 9.0 & 40.9 & 13.4 & 48.0 & 300\,/\,0\\
 & Prompt-to-Prompt~\cite{hertz2022prompt} & - & 52.1 & 47.3 & 46.9 & 37.5 & 5.0 & 9.0 & 44.2 & 12.4 & 48.2 & 200\,/\,0\\
 & Best-of-4 & 4 & 62.0 & 46.4 & 53.9 & 52.5 & 6.0 & 19.0 & 53.7 & 17.6 & 51.9 & 800\,/\,4\\
 & Best-of-8 & 8 & \textbf{65.7} & 52.3 & 56.3 & 57.5 & 7.0 & \textbf{20.0} & 59.2 & \textbf{20.0} & 55.1 & 1600\,/\,8\\
 & PSP~\cite{guimaraes2026psp} & 4 & 63.3 & 52.9 & 53.8 & 50.0 & 7.0 & 13.0 & 55.8 & 19.7 & 52.8 & 800\,/\,14\\
 & BFS~\cite{zhang2026classical} & 4 & 60.7 & 45.3 & 54.1 & 51.2 & 7.0 & 16.0 & 53.4 & 17.7 & 50.9 & 800\,/\,24\\
 & SVDD-PM~\cite{li2024svdd} & 4 & 55.8 & 44.1 & 47.5 & 32.5 & 2.0 & 14.0 & 50.3 & 13.8 & 49.4 & 794\,/\,396\\
 & LiDAR~\cite{kim2026lookahead} & 4 & 57.4 & 43.3 & 48.2 & 50.8 & 1.0 & 10.0 & 52.6 & 10.9 & 45.8 & 1300\,/\,54\\
 & DTS~\cite{jain2025dts} & 4 & 62.0 & 51.9 & 52.2 & 60.0 & 7.0 & 17.0 & 53.9 & 16.0 & 52.9 & 916\,/\,5\\
 & FK steering~\cite{singhal2025fk} & 4 & 59.5 & 46.2 & 53.0 & 52.5 & 5.0 & 15.0 & 51.4 & 17.8 & 51.6 & 800\,/\,24\\
 & \cellcolor{trailmixgreen} \methodname{} & \cellcolor{trailmixgreen} 4 & \cellcolor{trailmixgreen} 63.6 & \cellcolor{trailmixgreen} \textbf{57.6} & \cellcolor{trailmixgreen} \textbf{56.3} & \cellcolor{trailmixgreen} \textbf{61.3} & \cellcolor{trailmixgreen} \textbf{8.0} & \cellcolor{trailmixgreen} 19.0 & \cellcolor{trailmixgreen} \textbf{59.3} & \cellcolor{trailmixgreen} 19.2 & \cellcolor{trailmixgreen} \textbf{56.3} & \multicolumn{1}{>{\columncolor{trailmixgreen}[\tabcolsep][0pt]}c@{}}{928\,/\,12}\\
\midrule
\multirow{10}{*}{\rotatebox[origin=c]{90}{SDXL}} & Base model (single) & 1 & 69.6 & 55.4 & 54.0 & 40.0 & 6.0 & 19.0 & 54.5 & 19.6 & 52.4 & 200\,/\,1\\
\cmidrule(lr){2-13}
 & Best-of-4 & 4 & 76.8 & 62.7 & 63.0 & 53.7 & 9.0 & 38.0 & 64.0 & 21.6 & 57.6 & 800\,/\,4\\
 & Best-of-8 & 8 & 79.0 & 62.7 & 69.2 & \textbf{73.8} & 14.0 & \textbf{43.0} & 67.4 & 23.3 & 60.0 & 1600\,/\,8\\
 & PSP~\cite{guimaraes2026psp} & 4 & 78.0 & 63.3 & 65.2 & 61.3 & 16.0 & 37.0 & 66.9 & 24.6 & 58.5 & 800\,/\,14\\
 & BFS~\cite{zhang2026classical} & 4 & 76.1 & 60.0 & 63.6 & 60.0 & 12.0 & 32.0 & 64.2 & 24.5 & 57.8 & 800\,/\,24\\
 & SVDD-PM~\cite{li2024svdd} & 4 & 78.0 & 58.3 & 56.2 & 42.5 & 10.0 & 23.0 & 62.0 & 22.5 & 54.3 & 794\,/\,396\\
 & LiDAR~\cite{kim2026lookahead} & 4 & 75.0 & 53.0 & 59.5 & 60.8 & 12.0 & 29.0 & 60.2 & 19.2 & 51.2 & 1600\,/\,54\\
 & DTS & 4 & 79.3 & 60.6 & 63.4 & 65.0 & 11.0 & 30.0 & 65.2 & 22.8 & 56.2 & 916\,/\,5\\
 & FK steering~\cite{singhal2025fk} & 4 & 76.6 & 60.0 & 61.4 & 55.0 & 13.0 & 24.0 & 64.0 & 24.2 & 55.9 & 800\,/\,24\\
 & \cellcolor{trailmixgreen} \methodname{} & \cellcolor{trailmixgreen} 4 & \cellcolor{trailmixgreen} \textbf{80.2} & \cellcolor{trailmixgreen} \textbf{69.0} & \cellcolor{trailmixgreen} \textbf{69.4} & \cellcolor{trailmixgreen} 71.3 & \cellcolor{trailmixgreen} \textbf{17.0} & \cellcolor{trailmixgreen} 41.0 & \cellcolor{trailmixgreen} \textbf{68.4} & \cellcolor{trailmixgreen} \textbf{25.9} & \cellcolor{trailmixgreen} \textbf{60.7} & \multicolumn{1}{>{\columncolor{trailmixgreen}[\tabcolsep][0pt]}c@{}}{928\,/\,12}\\
\bottomrule
\end{tabular}
}
\caption{\textbf{Text-to-image generation.} Results are scored by independent benchmark evaluators, on a $0$-$100$ scale (higher is better). The final column reports U-Net network evaluations and judge calls per prompt. Bold indicates the highest score in each column for each model.}
\label{tab:t2i}
\end{table*}

\noindent
\textbf{Sampling and baselines.}
We generate images with $100$ stochastic DDIM steps ($\eta=1$)~\cite{song2021ddim} and share initial seeds for paired comparisons.
We compare \methodname{} with three groups of baselines.

\emph{(i) Independent sampling.}
The base model generates an image along a single denoising trajectory without reward-based selection. Best-of-$N$ generates $N$ independent images and selects the completed image with the highest reward. We evaluate Best-of-4 and Best-of-8.

\emph{(ii) Selection during denoising.}
These methods evaluate unfinished trajectories to decide where to allocate further computation. FK steering~\cite{singhal2025fk} periodically resamples trajectories using judge-derived weights, duplicating promising trajectories and discarding others. Resampling preserves the population size, although duplicated trajectories share an intermediate state and may produce similar images. We also compare with PSP~\cite{guimaraes2026psp}, BFS~\cite{zhang2026classical}, SVDD-PM~\cite{li2024svdd}, and DTS~\cite{jain2025dts}, which use alternative pruning, transition-selection, and search strategies.

\emph{(iii) Guidance and attention control.}
These methods modify generation within individual trajectories. LiDAR~\cite{kim2026lookahead}, Universal Guidance~\cite{bansal2023universalguidancediffusionmodels}, and FreeDoM~\cite{yu2023freedomtrainingfreeenergyguidedconditional} apply reward-derived corrections, while SEGA~\cite{brack2023segainstructingtexttoimagemodels} and Prompt-to-Prompt~\cite{hertz2022prompt} control semantic directions and attention, respectively.

We use four trajectories for \methodname{}, all selection-during-denoising baselines, and LiDAR. The remaining guidance and attention-control methods use a single trajectory, consistent with their typical implementations. We evaluate LiDAR on both models and adapt the other four methods to generation on SDv1.4. 

\vspace{2pt}
\noindent
\textbf{Steering reward.}
All reward-based methods use ImageReward~\cite{xu2023imagereward} to score prompt alignment and visual preference, except Universal Guidance and FreeDoM, which use CLIP-based energies. SEGA and Prompt-to-Prompt do not use an external reward judge.

\noindent
\textbf{Benchmarks and evaluation.}
We assess how well generated images satisfy their prompts on three benchmarks with independent evaluators, separate from the reward judges used during generation. Scores are reported on a $0$-$100$ scale, with higher values indicating better alignment. \textsc{RareBench}~\cite{park2025r2f} tests whether models can generate unusual objects and combinations of objects using $320$ prompts. GPT-4o~\cite{openai2024gpt4o} rates how well each image matches its prompt on a $1$-$5$ scale, which we rescale to $0$-$100$. We report single-object and multi-object results separately. \textsc{\dsGen}~\cite{ghosh2023geneval} evaluates $553$ prompts across six categories. According to the standard \dsGen{} protocol, we use Mask2Former \cite{cheng2022maskedattentionmasktransformeruniversal} to check object presence, counts, and spatial relationships, and CLIP ViT-L/14 \cite{pmlr-v139-radford21a} to check object colors and color attribution. We report the average success rate across all categories (Overall), along with Counting, Position, and Color attribution scores.\textsc{T2I-CompBench++}~\cite{huang2025t2icompbench} tests compositional alignment using $1{,}500$ prompts across five categories. Binding measures whether colors, shapes, and textures are assigned to the correct objects, averaging BLIP-VQA scores \cite{pmlr-v162-li22n} across these three attributes. UniDet \cite{Zhou_2022_CVPR} evaluates Spatial and Numeracy categories, verifying relationships between objects and object counts, respectively.

\noindent
\textbf{Inference cost.}
We strive for improved alignment while maintaining computational efficiency. We measure inference cost with two quantities: U-Net evaluations for denoising and judge calls for evaluating image alignment. Each denoiser prediction counts as one U-Net evaluation, so a $100$-step sample with classifier-free guidance requires $200$ evaluations. Judge calls count the number of candidate images scored by the reward model, including intermediate evaluations and lookahead previews. Both measures account for the additional computation required to generate and evaluate lookaheads. For Universal Guidance and FreeDoM, judge calls instead correspond to CLIP energy evaluations. We report these costs separately because their relative importance depends on the setting: denoising may dominate generation time, while scoring with a large reward model may make judging the bottleneck.

\begin{figure*}[!h]
    \centering
\includegraphics[width=0.78\textwidth]{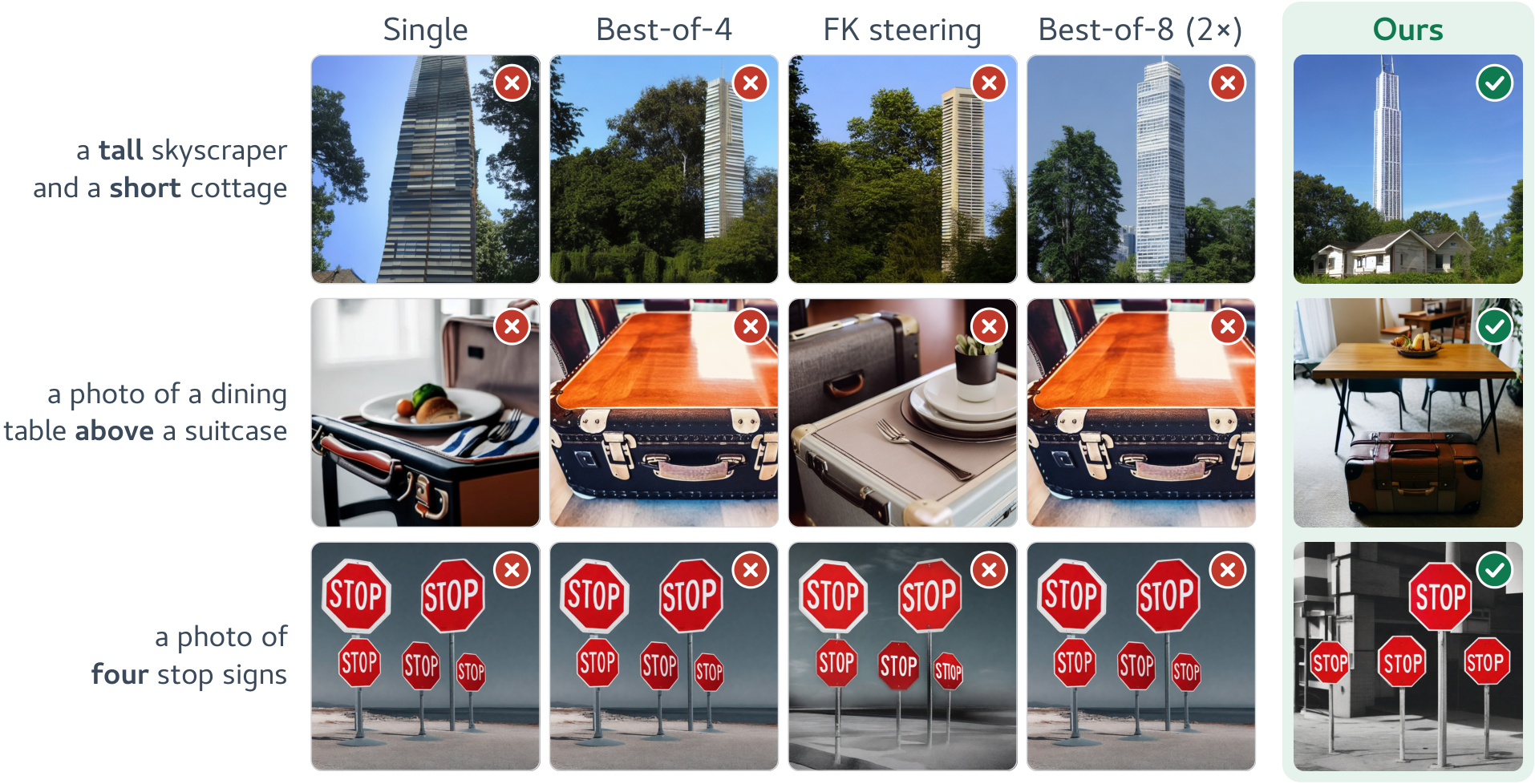}
    \caption{\textbf{Example outputs.}
    These examples illustrate how \methodname{} follows prompts with multiple objects, spatial relationships, and object counts compared to a few particle baselines. In particular, we can see how \methodname{} generates realistic outputs. }
    \label{fig:t2i-examples}
    \vspace{-10pt}
\end{figure*}

\subsubsection{Results}
\noindent
\textbf{\methodname{} improves alignment across semantic requirements, particularly for rare compositions.}
\methodname{} outperforms FK steering and LiDAR on every reported metric and matches or exceeds Best-of-4 across both diffusion models (\cref{tab:t2i}). It also achieves higher overall \dsGen{} scores than other trajectory selection and search methods, including PSP, BFS, SVDD-PM, and DTS. These improvements span object counting, spatial relationships, and attribute binding, with particularly strong results on \textsc{RareBench}. \methodname{} achieves the highest multi-object score among the evaluated methods on both models, surpassing even Best-of-8. Its gains over FK steering and LiDAR are also larger for multi-object compositions than for single objects. This suggests that combining lookahead evaluation with adaptive resampling is especially valuable when several uncommon concepts must appear together in one image (see examples in \cref{fig:t2i-examples}).

\vspace{4pt}
\noindent
\textbf{\methodname{} achieves strong alignment with fewer denoiser evaluations.}
\methodname{} matches or exceeds Best-of-8 on key alignment metrics with $42\%$ fewer U-Net evaluations, though it requires additional judge calls. Compared with FK steering, lookahead previews increase denoising work but halve judge calls, while \methodname{} achieves better alignment than LiDAR with lower costs on both measures. These comparisons suggest that how computation is allocated during generation matters, motivating a closer examination of performance as the particle budget increases.
\begin{figure}[!h]
    \centering \includegraphics[width=0.8\linewidth]{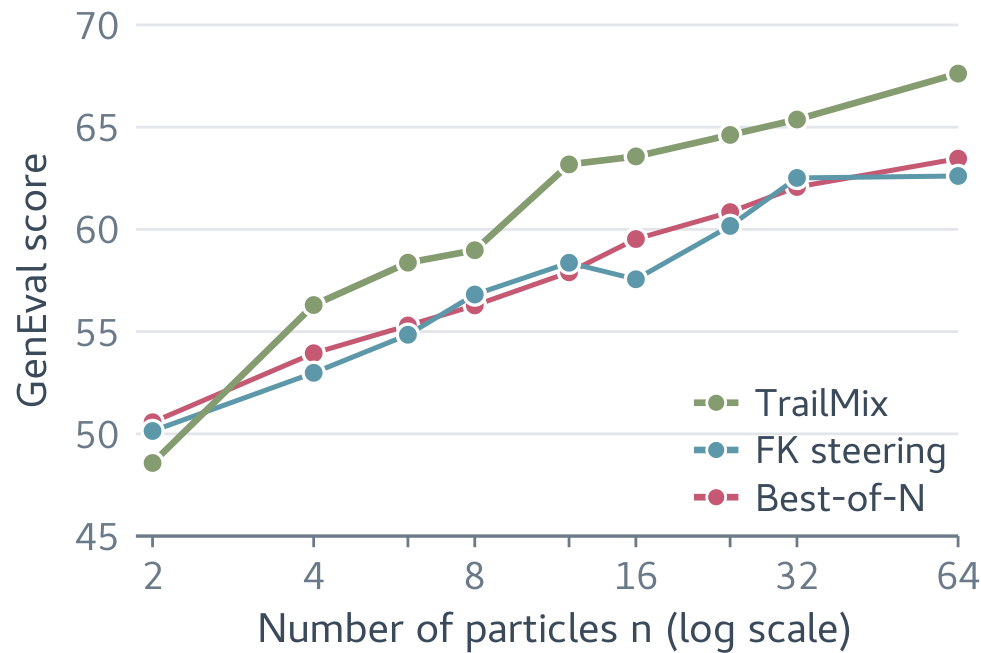}
    \caption{\textbf{Scaling with particle count.} \dsGen\ performance as the number of particles increases. \methodname{} leads FK steering and Best-of-$N$ at every tested count from $4$ to $64$.}
    \label{fig:particle_scaling}
    \vspace{-12pt}
\end{figure}

\noindent
\textbf{\methodname{} maintains its advantage as the particle budget grows.}
Increasing the number of particles generally improves generation quality, raising the question of whether the baselines can close the gap with larger populations. We evaluate this on GenEval by varying the particle budget for \methodname{}, FK steering, and Best-of-$N$ (\cref{fig:particle_scaling}). Although all three methods benefit from additional particles, \methodname{} achieves higher GenEval scores at every tested population size from $4$ to $64$. With only $16$ particles, it reaches approximately the same score as the baselines with $64$, and its advantage continues to grow at larger budgets. The only exception is $n=2$, where it underperforms both baselines. These results show that \methodname{} not only improves generation at a fixed particle budget, but also makes more effective use of larger populations.

\subsection{Image editing}
\label{sec:exp-edit}

\noindent
\textbf{Sampling and baselines.}
In this set of tasks, we evaluate how well methods change a requested attribute while preserving the rest of the image. This is challenging because models can often change unrelated objects or background details while prompted for a targeted change. We use stochastic DDIM sampling ($\eta=1$)~\cite{song2021ddim} and the same baselines as the text-to-image generation evaluation (\cref{sec:t2i}).

\begin{figure*}[!t]
\centering
\begin{tabular}{@{}p{0.38\textwidth}@{\hspace{0.02\textwidth}}p{0.60\textwidth}@{}}

\begin{minipage}[t]{\linewidth}
\vspace{0pt}
\centering
\includegraphics[width=\linewidth]{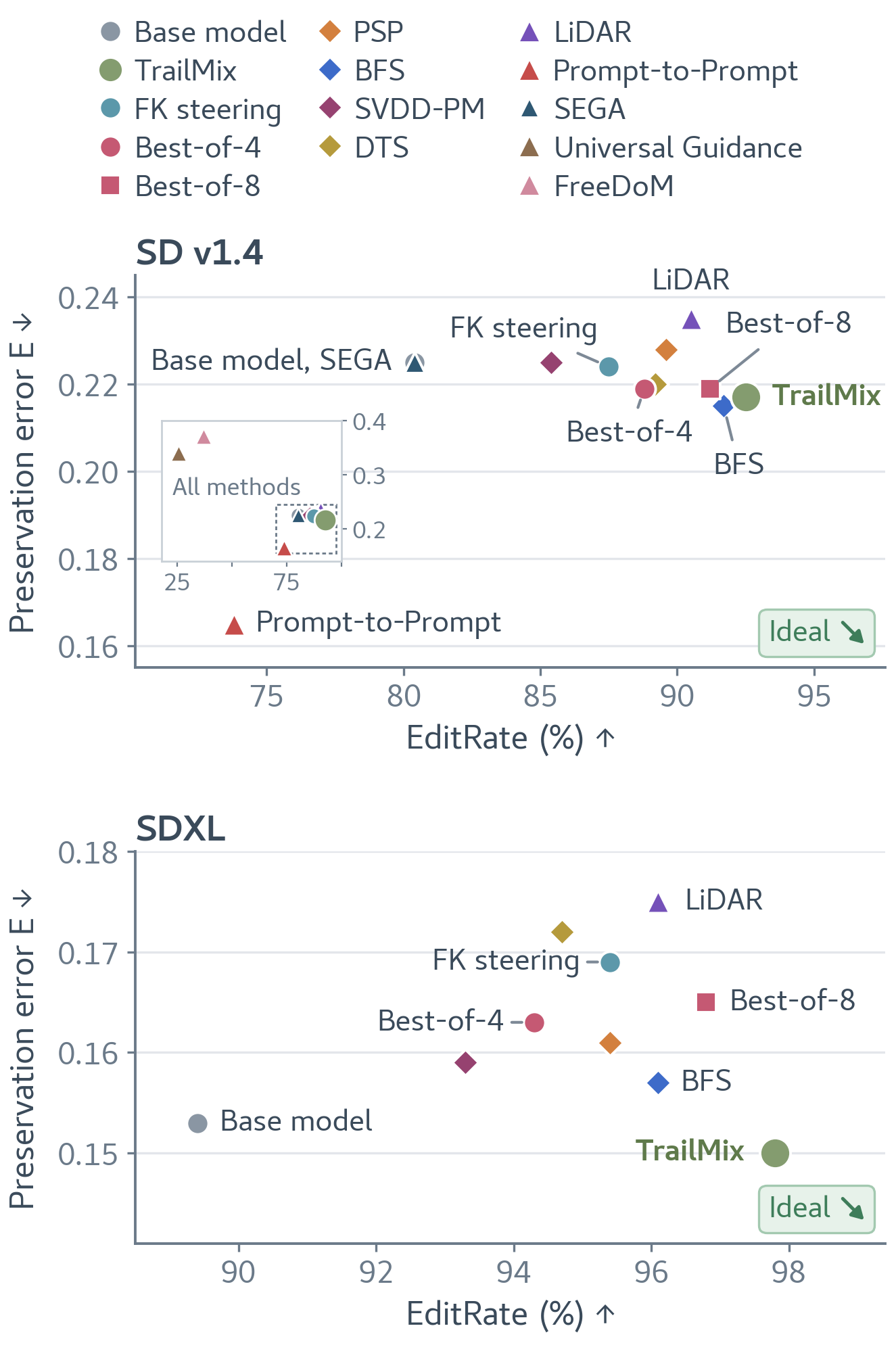}
\end{minipage}
&
\begin{minipage}[t]{\linewidth}
\vspace{0pt}
\centering
\footnotesize
\setlength{\tabcolsep}{3pt}
\renewcommand{\arraystretch}{1.12}
\resizebox{\linewidth}{!}{%
\begin{tabular}{@{}ll c cc ccccc@{}}
\toprule
& & & \multicolumn{2}{c}{\dsEdit{}} & \multicolumn{5}{c}{PIE-Bench}\\
\cmidrule(lr){4-5}\cmidrule(lr){6-10}
Model & Method & \shortstack{Particles\\$n$}
& $E_{\mathrm{ver}}\downarrow$
& \shortstack{Ctx.\\lost$\downarrow$}
& \shortstack{Edit\\Rate$\uparrow$}
& \shortstack{CLIP\\edit$\uparrow$}
& \shortstack{Struc-\\ture$\downarrow$}
& PSNR$\uparrow$ & LPIPS$\downarrow$\\
\midrule
\multirow{14}{*}{\rotatebox[origin=c]{90}{SD v1.4}}
& Base model (single) & 1 & 0.223 & 6.2 & 13.4 & 21.35 & 18.38 & 24.05 & 98.0\\
\cmidrule(lr){2-10}
& Universal Guidance~\cite{bansal2023universalguidancediffusionmodels} & - & 0.330 & 25.0 & 5.6 & 21.47 & \textbf{13.90} & \textbf{25.19} & \textbf{82.0}\\
& FreeDoM~\cite{yu2023freedomtrainingfreeenergyguidedconditional} & - & 0.384 & 16.2 & \textbf{29.8} & \textbf{21.85} & 135.41 & 15.40 & 335.4\\
& SEGA~\cite{brack2023segainstructingtexttoimagemodels} & - & 0.223 & 6.2 & 27.0 & 21.69 & 20.37 & 23.61 & 103.6\\
& Prompt-to-Prompt~\cite{hertz2022prompt} & - & \textbf{0.172} & 4.2 & 12.3 & 21.33 & 18.34 & 24.07 & 97.7\\
& Best-of-4 & 4 & 0.220 & 4.2 & 18.5 & 21.60 & 18.59 & 24.02 & 98.4\\
& Best-of-8 & 8 & 0.221 & \textbf{2.1} & 21.0 & 21.69 & 18.49 & 24.01 & 97.9\\
& PSP~\cite{guimaraes2026psp} & 4 & 0.223 & 3.8 & 21.6 & 21.71 & 18.44 & 24.03 & 98.1\\
& BFS~\cite{zhang2026classical} & 4 & 0.212 & 3.3 & 20.0 & 21.67 & 18.38 & 24.02 & 98.1\\
& SVDD-PM~\cite{li2024svdd} & 4 & 0.219 & 5.8 & 18.3 & 21.53 & 18.43 & 24.06 & 98.1\\
& LiDAR~\cite{kim2026lookahead} & 4 & 0.234 & 3.3 & 26.5 & 19.08 & 26.09 & 22.46 & 113.6\\
& DTS~\cite{jain2025dts} & 4 & 0.221 & 4.2 & 19.3 & 21.64 & 18.50 & 24.03 & 98.1\\
& FK steering~\cite{singhal2025fk} & 4 & 0.218 & 4.6 & 20.6 & 21.69 & 18.53 & 24.04 & 98.1\\
& \cellcolor{trailmixgreen} \methodname{} & \cellcolor{trailmixgreen} 4 & \cellcolor{trailmixgreen} 0.218 & \cellcolor{trailmixgreen} 2.9 & \cellcolor{trailmixgreen} 19.3 & \cellcolor{trailmixgreen} 21.65 & \cellcolor{trailmixgreen} 18.43 & \cellcolor{trailmixgreen} 24.08 & \multicolumn{1}{>{\columncolor{trailmixgreen}[\tabcolsep][0pt]}c@{}}{97.8}\\
\midrule
\multirow{10}{*}{\rotatebox[origin=c]{90}{SDXL}}
& Base model (single) & 1 & \textbf{0.150} & 0.7 & 8.2 & 20.94 & 14.40 & 26.36 & 84.8\\
\cmidrule(lr){2-10}
& Best-of-4 & 4 & 0.160 & \textbf{0.0} & 11.3 & 21.11 & 14.45 & 26.35 & 84.5\\
& Best-of-8 & 8 & 0.165 & 0.4 & 13.2 & 21.23 & 14.44 & 26.32 & 84.6\\
& PSP~\cite{guimaraes2026psp} & 4 & 0.158 & 0.7 & 12.3 & 21.18 & 14.45 & 26.37 & 84.7\\
& BFS~\cite{zhang2026classical} & 4 & 0.154 & 0.4 & 11.7 & 21.08 & 14.40 & 26.37 & 84.4\\
& SVDD-PM~\cite{li2024svdd} & 4 & 0.158 & 1.1 & 12.3 & 21.09 & 14.41 & 26.37 & 84.9\\
& LiDAR~\cite{kim2026lookahead} & 4 & 0.175 & 1.4 & \textbf{15.6} & 21.35 & 14.65 & 25.70 & 87.8\\
& DTS~\cite{jain2025dts} & 4 & 0.169 & 0.4 & 13.0 & 21.11 & \textbf{14.28} & 26.36 & 84.4\\
& FK steering~\cite{singhal2025fk} & 4 & 0.168 & \textbf{0.0} & 12.1 & 21.00 & 14.46 & 24.39 & 86.7\\
& \cellcolor{trailmixgreen} \methodname{} & \cellcolor{trailmixgreen} 4 & \cellcolor{trailmixgreen} 0.161 & \cellcolor{trailmixgreen} \textbf{0.0} & \cellcolor{trailmixgreen} 13.6 & \cellcolor{trailmixgreen} \textbf{22.08} & \cellcolor{trailmixgreen} 14.41 & \cellcolor{trailmixgreen} \textbf{26.38} & \multicolumn{1}{>{\columncolor{trailmixgreen}[\tabcolsep][0pt]}c@{}}{\textbf{84.1}}\\
\bottomrule
\end{tabular}%
}
\end{minipage}
\\[4pt]

\begin{minipage}[t]{\linewidth}
\vspace{0pt}
\captionof{figure}{\textbf{Editing tradeoff.}
Tradeoff between edit success and preservation of unrelated semantic details on \dsEdit{}, with higher EditRate and lower preservation error $E$ preferred.}
\label{fig:edit-tradeoff}
\end{minipage}
&
\begin{minipage}[t]{\linewidth}
\vspace{0pt}
\captionof{table}{\textbf{Additional image editing results.}
Additional metrics on \dsEdit{} and editing results on PIE-Bench.
EditRate and Context lost are percentages; Structure and LPIPS are multiplied by $10^3$.
Arrows indicate preferred directions; bold marks the best result, including ties.}
\label{tab:edit}
\end{minipage}
\\
\end{tabular}
\vspace{-12pt}
\end{figure*}

\vspace{2pt}
\noindent
\textbf{Steering reward.}
We use Qwen2.5-VL-7B~\cite{bai2025qwen25vl} to score whether candidate edits contain the target attribute and no longer contain the original attribute. Best-of-$N$ uses these scores to select a completed edit. \methodname{} and the resampling and search baselines use them to select trajectories during denoising, while LiDAR uses lookahead scores to guide individual trajectories. Universal Guidance and FreeDoM use CLIP-based energies; SEGA and Prompt-to-Prompt do not use an external reward judge.

\vspace{2pt}
\noindent
\textbf{Benchmarks and evaluation.}
Existing image-editing benchmarks such as PIE-Bench evaluate edit success and source preservation using text-image alignment and visual similarity metrics~\cite{ju2024pnp}. We introduce \dsEdit{} (Semantic Editing and Attribute Maintenance Benchmark) to explicitly assess whether changing a requested attribute preserves other attributes in the image. The benchmark contains $316$ controlled source-target description pairs differing in a single attribute, spanning $12$ edit types including color, action, and material. For each pair, we generate questions to evaluate both the requested change and preservation of unrelated attributes. Because diffusion models can learn stereotypical associations~\cite{luccioni2023stablebiasanalyzingsocietal,friedrich2023fairdiffusion}, we also include a separate cohort of $192$ edits toward and away from these associations to assess counter-stereotypical editing and directional asymmetry. Source images are generated using the evaluated diffusion model, allowing us to reuse their initial noise during editing. Further details are provided in App.~\ref{sec:supp-seam-bench}.

We additionally evaluate on \textsc{PIE-Bench}~\cite{ju2024pnp}, which provides $700$ existing source images with editing instructions and masks. Because their initial noise is unknown, we use SDEdit~\cite{meng2022sdedit} to initialize editing from a noised version of each source. We use a strength of $s=0.3$, selected using the base model and fixed across both backbones.

For both benchmarks, we use Gemma-3-27B~\cite{gemma2025gemma3} to evaluate edit success independently of the steering judge. We first check that each source image contains the intended subject and original attribute, but not the target attribute, ensuring that the requested edit requires an actual change. An edit succeeds when the output retains the subject, removes the original attribute, and introduces the target attribute. EditRate is the percentage of these source images successfully edited, while Context lost measures how often the subject disappears.

We measure preservation separately from edit success. On \dsEdit{}, Gemma evaluates attributes that should remain unchanged. Preservation error $E$ is the average absolute change in its confidence that these attributes are present between the source and edited images. We also report $E_{\mathrm{verified}}$, which measures the same error on edits where the attribute was successfully edited. Lower values indicate better preservation. On PIE-Bench, the supplied masks identify the regions intended to change. We measure target-prompt alignment within these regions using $\mathrm{CLIP}_{\mathrm{edit}}$ and preservation outside them using PSNR and LPIPS. PSNR compares pixel values, while LPIPS measures perceptual differences using neural network features. Higher PSNR and lower LPIPS indicate better preservation. Structure measures changes across the whole image using DINO features; lower values are better.

For counter-stereotypical edits, \emph{Against} measures success when editing away from stereotypical attributes. \emph{Asymmetry} is the absolute difference in success rates between the two editing directions, measured only for contexts where the source images meet the starting conditions for both directions.

\subsubsection{Results}
\label{sec:edit-results}

\noindent
\textbf{\methodname{} improves edit success while preserving source content.}
On \dsEdit{}, \methodname{} achieves the highest EditRate on both backbones while maintaining low preservation error (\cref{fig:edit-tradeoff}). On SDv1.4, it exceeds LiDAR in edit success (92.5\% versus 90.5\%) while reducing overall preservation error and subject loss. On SDXL, it attains 97.8\% EditRate with the lowest overall preservation error and no observed subject loss, although the base model has lower preservation error when considering only successful edits (\cref{tab:edit}). These results show that higher edit success can be achieved without substantially altering unrelated source content.

\vspace{2pt}
\noindent
\textbf{\methodname{} balances edit alignment and preservation on existing images.}
On PIE-Bench, \methodname{} achieves the highest edited-region text alignment and lowest LPIPS on SDXL, along with the highest EditRate among particle-based selection baselines (\cref{tab:edit}). LiDAR achieves higher edit success (15.6\% versus 13.6\%) but introduces larger structural and perceptual changes. On SDv1.4, \methodname{} improves EditRate over the base model and Best-of-4 and retains strong source preservation, although several baselines achieve higher edit success. Thus, the SDXL results offer the clearest evidence of improved edit alignment with limited changes to the source.

\begin{table}[!h]
\centering
\footnotesize
\setlength{\tabcolsep}{3pt}
\begin{tabular}{@{}lccccc@{}}
\toprule
& & \multicolumn{2}{c}{SD v1.4} & \multicolumn{2}{c}{SDXL}\\
\cmidrule(lr){3-4}\cmidrule(lr){5-6}
Method & \shortstack{Particles\\$n$} & Against$\uparrow$ & Asym.$\downarrow$ & Against$\uparrow$ & Asym.$\downarrow$\\
\midrule
Single & 1 & 78.8 & 5.3 & 76.7 & 4.3\\
Best-of-4 & 4 & 82.5 & \textbf{1.8} & 83.3 & 5.8\\
Best-of-8 & 8 & \textbf{85.0} & 7.0 & 83.3 & 4.3\\
FK steering~\cite{singhal2025fk} & 4 & 77.5 & 5.3 & 81.1 & 5.8\\
\rowcolor{trailmixgreen}
\methodname{} & 4 & 83.8 & \textbf{1.8} & \textbf{90.0} & \textbf{1.4}\\
\bottomrule
\end{tabular}
\caption{\textbf{Counter-stereotypical editing.} Results are reported as percentages; bold shows best result per column, including ties.}
\label{tab:bias}
\vspace{-10pt}
\end{table}

\begin{figure}[!h]
    \centering
    \includegraphics[width=0.8\linewidth]{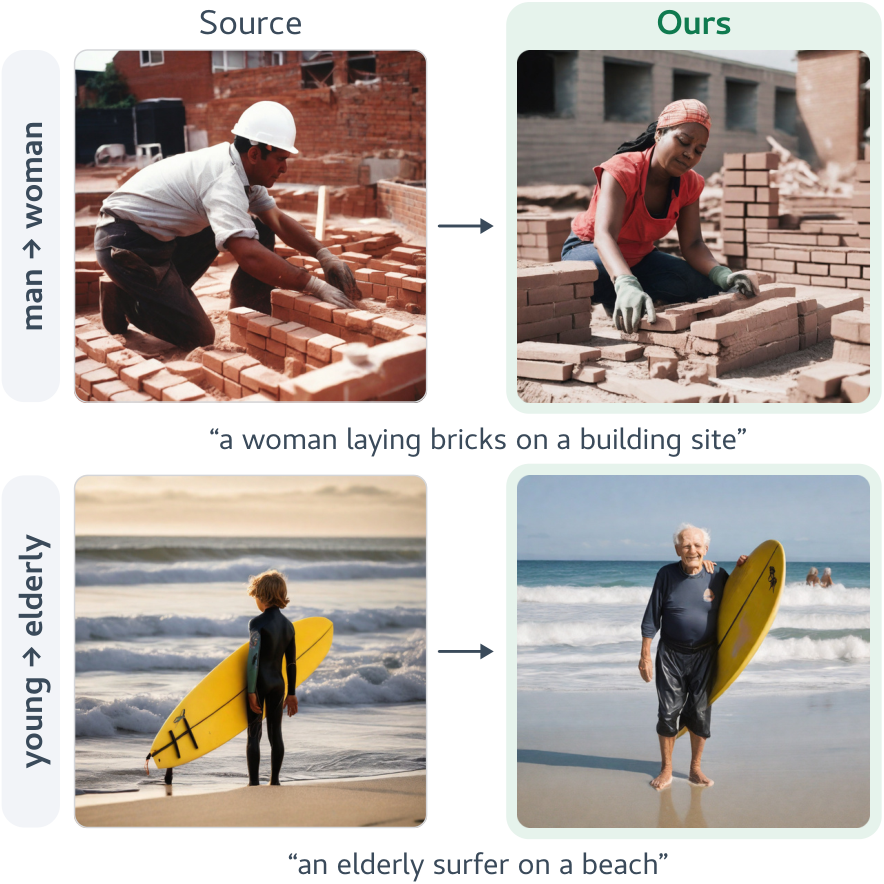}
    \caption{\textbf{Counter-stereotypical image editing.} Source images (left) and \methodname{} outputs (right).}
    \label{fig:counterstereotypical-edits}
    \vspace{-10pt}
\end{figure}

\noindent
\textbf{\methodname{} improves counter-stereotypical editing while reducing asymmetry.}
\methodname{} achieves the lowest \emph{Asymmetry} on both models, tying Best-of-4 on SDv1.4 (\cref{tab:bias}). On SDXL, it reaches 90.0\% counter-stereotypical EditRate, compared with 83.3\% for the strongest baseline, while reducing Asymmetry to 1.4 percentage points. On SDv1.4, Best-of-8 achieves higher counter-stereotypical edit success but substantially greater asymmetry. These results suggest that \methodname{} helps counter stereotypical associations while balancing success across editing directions. The examples in \cref{fig:counterstereotypical-edits} illustrate edits to the bricklayer's gender and the surfer's age while retaining their activities and scene context.

\subsection{Ablations}
\label{sec:ablations}

We examine the two main components of \methodname{}: lookahead-based reward estimation and adaptive steering strength. We vary the lookahead length and compare adaptive and fixed steering strengths on SDv1.4 with four particles, holding all other settings fixed (\cref{tab:abl-summary}). We perform additional ablations to investigate the impact of reward threshold, preview budget, particle count, resampling schedule, and reward judges (App. \ref{sec:supp-ablations}).

\vspace{5pt}
\noindent
\textbf{Short lookaheads improve alignment relative to one-step predictions.}
Four- and eight-step previews outperform one-step predictions on both \dsGen{} and \textsc{RareBench}, with the largest gains on \textsc{RareBench}. Extending previews to sixteen steps increases computation without improving performance over eight steps. As further analyzed in App. \ref{sec:supp-rank}, lookahead previews are particularly useful early in denoising, as they can more reliably identify trajectories likely to produce successful images than one-step predictions.

\vspace{3pt}
\noindent
\textbf{Adaptive steering adjusts selection pressure across prompts.}
Different fixed steering strengths perform best on \dsGen{} and \textsc{RareBench}, while adaptive steering matches the best tested setting on both without benchmark-specific tuning. Further analysis shows that adaptive steering generally applies weaker selection when the base model already produces high-reward images, retaining more distinct initial trajectories than fixed steering while maintaining competitive alignment (App. \ref{sec:supp-strength}).
\begin{table}[!t]
\centering
\footnotesize
\setlength{\tabcolsep}{2.5pt}
\renewcommand{\arraystretch}{1.1}
\begin{tabular}{@{}lccc@{}}
\toprule
Variant & \dsGen & \textsc{RareBench} & \shortstack{U-Net /\\judge calls}\\
\midrule
\multicolumn{4}{@{}l}{\emph{Lookahead length}}\\
$L=1$ & 54.4 & 57.0 & 816\,/\,12\\
$L=4$ & 54.8 & 61.1 & 864\,/\,12\\
$L=8$ (\methodname{}) & \textbf{56.3} & \textbf{61.4} & 928\,/\,12\\
$L=16$ & 55.6 & 58.9 & 1056\,/\,12\\
\midrule
\multicolumn{4}{@{}l}{\emph{Steering strength}}\\
Fixed $\beta=1$ & 55.2 & \textbf{61.4} & 928\,/\,12\\
Fixed $\beta=5$ & 54.9 & 60.5 & 928\,/\,12\\
Fixed $\beta=20$ & \textbf{56.3} & 60.0 & 928\,/\,12\\
Adaptive (\methodname{}) & \textbf{56.3} & \textbf{61.4} & 928\,/\,12\\
\bottomrule
\end{tabular}
\caption{\textbf{Lookahead and steering-strength ablations.} All runs use SDv1.4 with four particles. Scores use independent benchmark evaluators on a $0$-$100$ scale. $L=1$ is a one-step prediction. Costs are per prompt and include previews. Bold marks the best score within each block, including ties.}
\label{tab:abl-summary}
\vspace{-10pt}
\end{table}

\section{Conclusion}
\label{sec:conclusion}

We introduce \methodname{}, a training-free approach to diffusion steering that uses the pretrained model's possible outcomes to improve text-image alignment. By evaluating inexpensive lookahead previews, \methodname{} identifies promising trajectories and adapts how strongly to favor them during resampling. Across two diffusion models, this approach improves compositional generation, with particularly strong gains on rare multi-object scenes. We further demonstrate its applicability to targeted image editing and introduce \dsEdit{}, a systematic benchmark for evaluating semantic edit success and source preservation, including edits against stereotypical associations. Together, our results show how adaptive reward estimation and resampling can better leverage the pretrained generative distribution without additional training.
\section*{LLM Usage Statement}

We used GPT-5.6 Luna to assist with identifying select baselines to ensure proper and timely empirical breadth. Claude Code Opus was used to assist with debugging, implementing baseline methods, and running experimental grids. All LLM-assisted outputs were reviewed and verified by the authors.

{
    \small
    \bibliographystyle{ieeenat_fullname}
    \bibliography{main}

@String(PAMI  = {IEEE TPAMI})

@String(CVPR  = {CVPR})

@String(ICCV  = {ICCV})

@String(NIPS  = {NeurIPS})

@String(ICLR  = {ICLR})

@inproceedings{sargent2026vlic,
  title={VLIC: Vision-Language Models As Perceptual Judges for Human-Aligned Image Compression},
  author={Sargent, Kyle and Gao, Ruiqi and Henzler, Philipp and Herrmann, Charles and Holynski, Aleksander and Fei-Fei, Li and Wu, Jiajun and Zhang, Jason Y},
  booktitle={Proceedings of the IEEE/CVF Conference on Computer Vision and Pattern Recognition},
  pages={10462--10471},
  year={2026}
}

@inproceedings{park2025r2f,
  title={Rare-to-Frequent: Unlocking Compositional Generation Power of Diffusion Models on Rare Concepts with {LLM} Guidance},
  author={Park, Dongmin and Kim, Sebin and Moon, Taehong and Kim, Minkyu and Lee, Kangwook and Cho, Jaewoong},
  booktitle=ICLR,
  pages={14650--14676},
  year={2025},
  url={https://proceedings.iclr.cc/paper_files/paper/2025/file/262ba7d20578417435414ce00e69fb12-Paper-Conference.pdf}
}

@article{friedrich2023fairdiffusion,
  title={Fair Diffusion: Instructing Text-to-Image Generation Models on Fairness},
  author={Friedrich, Felix and Brack, Manuel and Struppek, Lukas and Hintersdorf, Dominik and Schramowski, Patrick and Luccioni, Sasha and Kersting, Kristian},
  journal={arXiv preprint arXiv:2302.10893},
  year={2023}
}

@article{dong2024continually,
  title={How to continually adapt text-to-image diffusion models for flexible customization?},
  author={Dong, Jiahua and Liang, Wenqi and Li, Hongliu and Zhang, Duzhen and Cao, Meng and Ding, Henghui and Khan, Salman and Khan, Fahad S},
  journal={Advances in Neural Information Processing Systems},
  volume={37},
  pages={130057--130083},
  year={2024}
}

@misc{bansal2023universalguidancediffusionmodels,
      title={Universal Guidance for Diffusion Models}, 
      author={Arpit Bansal and Hong-Min Chu and Avi Schwarzschild and Soumyadip Sengupta and Micah Goldblum and Jonas Geiping and Tom Goldstein},
      year={2023},
      eprint={2302.07121},
      archivePrefix={arXiv},
      primaryClass={cs.CV},
      url={https://arxiv.org/abs/2302.07121}, 
}

@misc{guo2026earlyqualityassessmenttexttoimage,
      title={Toward Early Quality Assessment of Text-to-Image Diffusion Models}, 
      author={Huanlei Guo and Hongxin Wei and Bingyi Jing},
      year={2026},
      eprint={2603.02829},
      archivePrefix={arXiv},
      primaryClass={cs.CV},
      url={https://arxiv.org/abs/2603.02829}, 
}

@misc{yue2024exploringdiffusiontimestepsunsupervised,
      title={Exploring Diffusion Time-steps for Unsupervised Representation Learning}, 
      author={Zhongqi Yue and Jiankun Wang and Qianru Sun and Lei Ji and Eric I-Chao Chang and Hanwang Zhang},
      year={2024},
      eprint={2401.11430},
      archivePrefix={arXiv},
      primaryClass={cs.CV},
      url={https://arxiv.org/abs/2401.11430}, 
}

@misc{tinaz2026emergenceevolutioninterpretableconcepts,
      title={Emergence and Evolution of Interpretable Concepts in Diffusion Models}, 
      author={Berk Tinaz and Zalan Fabian and Mahdi Soltanolkotabi},
      year={2026},
      eprint={2504.15473},
      archivePrefix={arXiv},
      primaryClass={cs.CV},
      url={https://arxiv.org/abs/2504.15473}, 
}

@article{chatterjee2018sample,
  title={The sample size required in importance sampling},
  author={Chatterjee, Sourav and Diaconis, Persi},
  journal={The Annals of Applied Probability},
  volume={28},
  number={2},
  pages={1099--1135},
  year={2018},
  publisher={JSTOR}
}

@inproceedings{zhang2023itigen,
  title={{ITI-GEN}: Inclusive Text-to-Image Generation},
  author={Zhang, Cheng and Chen, Xuanbai and Chai, Siqi and Wu, Chen Henry and Lagun, Dmitry and Beeler, Thabo and De la Torre, Fernando},
  booktitle=ICCV,
  year={2023}
}

@inproceedings{shen2024finetuning,
  title={Finetuning Text-to-Image Diffusion Models for Fairness},
  author={Shen, Xudong and Du, Chao and Pang, Tianyu and Lin, Min and Wong, Yongkang and Kankanhalli, Mohan},
  booktitle=ICLR,
  year={2024}
}

@inproceedings{ma2025inferencescaling,
  title={Scaling Inference Time Compute for Diffusion Models},
  author={Ma, Nanye and Tong, Shangyuan and Jia, Haolin and others},
  booktitle=CVPR,
  year={2025}
}

@article{guimaraes2026psp,
  title={Inference-Time Scaling of Diffusion Models via Progressive Seed Pruning},
  author={Guimaraes, Rogerio and Perona, Pietro},
  journal={arXiv preprint arXiv:2607.21591},
  year={2026}
}

@inproceedings{singhal2025fk,
  title={A General Framework for Inference-time Scaling and Steering of Diffusion Models},
  author={Singhal, Raghav and Horvitz, Zachary and Teehan, Ryan and Ren, Mengye and Yu, Zhou and McKeown, Kathleen and Ranganath, Rajesh},
  booktitle={Int. Conf. Mach. Learn.},
  year={2025}
}

@inproceedings{wu2023tds,
  title={Practical and Asymptotically Exact Conditional Sampling in Diffusion Models},
  author={Wu, Luhuan and Trippe, Brian L. and Naesseth, Christian A. and Blei, David M. and Cunningham, John P.},
  booktitle=NIPS,
  year={2023}
}

@article{li2024svdd,
  title={Derivative-Free Guidance in Continuous and Discrete Diffusion Models with Soft Value-Based Decoding},
  author={Li, Xiner and Zhao, Yulai and Wang, Chenyu and Scalia, Gabriele and Eraslan, Gokcen and Nair, Surag and Biancalani, Tommaso and Ji, Shuiwang and Regev, Aviv and Levine, Sergey and Uehara, Masatoshi},
  journal={arXiv preprint arXiv:2408.08252},
  year={2024}
}

@inproceedings{rombach2022ldm,
  title={High-Resolution Image Synthesis with Latent Diffusion Models},
  author={Rombach, Robin and Blattmann, Andreas and Lorenz, Dominik and Esser, Patrick and Ommer, Bj{\"o}rn},
  booktitle=CVPR,
  year={2022}
}

@inproceedings{song2021ddim,
  title={Denoising Diffusion Implicit Models},
  author={Song, Jiaming and Meng, Chenlin and Ermon, Stefano},
  booktitle=ICLR,
  year={2021}
}

@inproceedings{ghosh2023geneval,
  title={{GenEval}: An Object-Focused Framework for Evaluating Text-to-Image Alignment},
  author={Ghosh, Dhruba and Hajishirzi, Hanna and Schmidt, Ludwig},
  booktitle=NIPS,
  year={2023}
}

@inproceedings{xu2023imagereward,
  title={{ImageReward}: Learning and Evaluating Human Preferences for Text-to-Image Generation},
  author={Xu, Jiazheng and Liu, Xiao and Wu, Yuchen and Tong, Yuxuan and Li, Qinkai and Ding, Ming and Tang, Jie and Dong, Yuxiao},
  booktitle=NIPS,
  year={2023}
}

@inproceedings{meng2022sdedit,
  title={{SDEdit}: Guided Image Synthesis and Editing with Stochastic Differential Equations},
  author={Meng, Chenlin and He, Yutong and Song, Yang and Song, Jiaming and Wu, Jiajun and Zhu, Jun-Yan and Ermon, Stefano},
  booktitle=ICLR,
  year={2022}
}

@inproceedings{ju2024pnp,
  title={{PnP} Inversion: Boosting Diffusion-based Editing with 3 Lines of Code},
  author={Ju, Xuan and Zeng, Ailing and Bian, Yuxuan and Liu, Shaoteng and Xu, Qiang},
  booktitle=ICLR,
  year={2024}
}

@article{bai2025qwen25vl,
  title={{Qwen2.5-VL} Technical Report},
  author={Bai, Shuai and others},
  journal={arXiv preprint arXiv:2502.13923},
  year={2025}
}

@article{gemma2025gemma3,
  title={Gemma 3 Technical Report},
  author={{Gemma Team}},
  journal={arXiv preprint arXiv:2503.19786},
  year={2025}
}

@inproceedings{kirstain2023pickapic,
  title={Pick-a-Pic: An Open Dataset of User Preferences for Text-to-Image Generation},
  author={Kirstain, Yuval and Polyak, Adam and Singer, Uriel and Matiana, Shahbuland and Penna, Joe and Levy, Omer},
  booktitle=NIPS,
  year={2023}
}

@article{huang2025t2icompbench,
  title={{T2I-CompBench++}: An Enhanced and Comprehensive Benchmark for Compositional Text-to-Image Generation},
  author={Huang, Kaiyi and Duan, Chengqi and Sun, Kaiyue and Xie, Enze and Li, Zhenguo and Liu, Xihui},
  journal=PAMI,
  year={2025}
}

@article{openai2024gpt4o,
  title={{GPT-4o} System Card},
  author={{OpenAI}},
  journal={arXiv preprint arXiv:2410.21276},
  year={2024}
}

@inproceedings{tumanyan2022splicing,
  title={Splicing {ViT} Features for Semantic Appearance Transfer},
  author={Tumanyan, Narek and Bar-Tal, Omer and Bagon, Shai and Dekel, Tali},
  booktitle=CVPR,
  year={2022}
}

@inproceedings{podell2024sdxl,
  title={{SDXL}: Improving Latent Diffusion Models for High-Resolution Image Synthesis},
  author={Podell, Dustin and English, Zion and Lacey, Kyle and Blattmann, Andreas and Dockhorn, Tim and M{\"u}ller, Jonas and Penna, Joe and Rombach, Robin},
  booktitle=ICLR,
  year={2024}
}

@inproceedings{kim2025das,
  title={Test-time Alignment of Diffusion Models without Reward Over-optimization},
  author={Kim, Sunwoo and Kim, Minkyu and Park, Dongmin},
  booktitle=ICLR,
  year={2025}
}

@inproceedings{jain2025dts,
  title={Diffusion Tree Sampling: Scalable Inference-Time Alignment of Diffusion Models},
  author={Jain, Vineet and Sareen, Kusha and Pedramfar, Mohammad and Ravanbakhsh, Siamak},
  booktitle=NIPS,
  year={2025}
}

@article{lamba2025alignment,
  title={Alignment and safety of diffusion models via reinforcement learning and reward modeling: A survey},
  author={Lamba, Preeti and Ravish, Kiran and Kushwaha, Ankita and Kumar, Pawan},
  journal={arXiv preprint arXiv:2505.17352},
  year={2025}
}

@misc{luccioni2023stablebiasanalyzingsocietal,
      title={Stable Bias: Analyzing Societal Representations in Diffusion Models}, 
      author={Alexandra Sasha Luccioni and Christopher Akiki and Margaret Mitchell and Yacine Jernite},
      year={2023},
      eprint={2303.11408},
      archivePrefix={arXiv},
      primaryClass={cs.CY},
      url={https://arxiv.org/abs/2303.11408}, 
}

@inproceedings{jena2025elucidating,
  title={Elucidating optimal reward-diversity tradeoffs in text-to-image diffusion models},
  author={Jena, Rohit and Taghibakhshi, Ali and Jain, Sahil and Shen, Gerald and Tajbakhsh, Nima and Vahdat, Arash},
  booktitle={2025 IEEE/CVF Winter Conference on Applications of Computer Vision (WACV)},
  pages={232--242},
  year={2025},
  organization={IEEE}
}

@InProceedings{Zhou_2022_CVPR,
    author    = {Zhou, Xingyi and Koltun, Vladlen and Kr\"ahenb\"uhl, Philipp},
    title     = {Simple Multi-Dataset Detection},
    booktitle = {Proceedings of the IEEE/CVF Conference on Computer Vision and Pattern Recognition (CVPR)},
    month     = {June},
    year      = {2022},
    pages     = {7571-7580}
}

@InProceedings{pmlr-v162-li22n,
  title = 	 {{BLIP}: Bootstrapping Language-Image Pre-training for Unified Vision-Language Understanding and Generation},
  author =       {Li, Junnan and Li, Dongxu and Xiong, Caiming and Hoi, Steven},
  booktitle = 	 {Proceedings of the 39th International Conference on Machine Learning},
  pages = 	 {12888--12900},
  year = 	 {2022},
  editor = 	 {Chaudhuri, Kamalika and Jegelka, Stefanie and Song, Le and Szepesvari, Csaba and Niu, Gang and Sabato, Sivan},
  volume = 	 {162},
  series = 	 {Proceedings of Machine Learning Research},
  month = 	 {17--23 Jul},
  publisher =    {PMLR},
  url = 	 {https://proceedings.mlr.press/v162/li22n.html}
}

@InProceedings{pmlr-v139-radford21a,
  title = 	 {Learning Transferable Visual Models From Natural Language Supervision},
  author =       {Radford, Alec and Kim, Jong Wook and Hallacy, Chris and Ramesh, Aditya and Goh, Gabriel and Agarwal, Sandhini and Sastry, Girish and Askell, Amanda and Mishkin, Pamela and Clark, Jack and Krueger, Gretchen and Sutskever, Ilya},
  booktitle = 	 {Proceedings of the 38th International Conference on Machine Learning},
  pages = 	 {8748--8763},
  year = 	 {2021},
  editor = 	 {Meila, Marina and Zhang, Tong},
  volume = 	 {139},
  series = 	 {Proceedings of Machine Learning Research},
  month = 	 {18--24 Jul},
  publisher =    {PMLR},
  url = 	 {https://proceedings.mlr.press/v139/radford21a.html}
}

@misc{cheng2022maskedattentionmasktransformeruniversal,
      title={Masked-attention Mask Transformer for Universal Image Segmentation}, 
      author={Bowen Cheng and Ishan Misra and Alexander G. Schwing and Alexander Kirillov and Rohit Girdhar},
      year={2022},
      eprint={2112.01527},
      archivePrefix={arXiv},
      primaryClass={cs.CV},
      url={https://arxiv.org/abs/2112.01527}, 
}

@misc{brack2023segainstructingtexttoimagemodels,
      title={SEGA: Instructing Text-to-Image Models using Semantic Guidance}, 
      author={Manuel Brack and Felix Friedrich and Dominik Hintersdorf and Lukas Struppek and Patrick Schramowski and Kristian Kersting},
      year={2023},
      eprint={2301.12247},
      archivePrefix={arXiv},
      primaryClass={cs.CV},
      url={https://arxiv.org/abs/2301.12247}, 
}

@misc{yu2023freedomtrainingfreeenergyguidedconditional,
      title={FreeDoM: Training-Free Energy-Guided Conditional Diffusion Model}, 
      author={Jiwen Yu and Yinhuai Wang and Chen Zhao and Bernard Ghanem and Jian Zhang},
      year={2023},
      eprint={2303.09833},
      archivePrefix={arXiv},
      primaryClass={cs.CV},
      url={https://arxiv.org/abs/2303.09833}, 
}

@inproceedings{zhang2026classical,
  title={Inference-time Scaling of Diffusion Models through Classical Search},
  author={Zhang, Xiangcheng and Lin, Haowei and Ye, Haotian and Zou, James and Ma, Jianzhu and Liang, Yitao and Du, Yilun},
  booktitle=ICLR,
  year={2026}
}

@inproceedings{black2024ddpo,
  title={Training Diffusion Models with Reinforcement Learning},
  author={Black, Kevin and Janner, Michael and Du, Yilun and Kostrikov, Ilya and Levine, Sergey},
  booktitle=ICLR,
  year={2024}
}

@inproceedings{fan2023dpok,
  title={{DPOK}: Reinforcement Learning for Fine-tuning Text-to-Image Diffusion Models},
  author={Fan, Ying and Watkins, Olivia and Du, Yuqing and Liu, Hao and Ryu, Moonkyung and Boutilier, Craig and Abbeel, Pieter and Ghavamzadeh, Mohammad and Lee, Kangwook and Lee, Kimin},
  booktitle=NIPS,
  year={2023}
}

@inproceedings{kim2026lookahead,
  title={Lookahead Sample Reward Guidance for Test-Time Scaling of Diffusion Models},
  author={Kim, Yeongmin and Shin, Donghyeok and Na, Byeonghu and Park, Minsang and Kim, Richard Lee and Moon, Il-Chul},
  booktitle={Proceedings of the International Conference on Machine Learning},
  series={Proceedings of Machine Learning Research},
  volume={306},
  pages={58969--58995},
  year={2026},
  url={https://proceedings.mlr.press/v306/kim26bo.html}
}

@article{zarei2024improving,
  title={Improving compositional attribute binding in text-to-image generative models via enhanced text embeddings},
  author={Zarei, Arman and Rezaei, Keivan and Basu, Samyadeep and Saberi, Mehrdad and Moayeri, Mazda and Kattakinda, Priyatham and Feizi, Soheil},
  journal={arXiv preprint arXiv:2406.07844},
  year={2024}
}

@inproceedings{clark2024draft,
  title={Directly Fine-Tuning Diffusion Models on Differentiable Rewards},
  author={Clark, Kevin and Vicol, Paul and Swersky, Kevin and Fleet, David J.},
  booktitle=ICLR,
  year={2024}
}

@inproceedings{wallace2024diffusiondpo,
  title={Diffusion Model Alignment Using Direct Preference Optimization},
  author={Wallace, Bram and Dang, Meihua and Rafailov, Rafael and Zhou, Linqi and Lou, Aaron and Purushwalkam, Senthil and Ermon, Stefano and Xiong, Caiming and Joty, Shafiq and Naik, Nikhil},
  booktitle=CVPR,
  year={2024}
}

@inproceedings{hertz2022prompt,
  title={Prompt-to-Prompt Image Editing with Cross-Attention Control},
  author={Hertz, Amir and Mokady, Ron and Tenenbaum, Jay and Aberman, Kfir and Pritch, Yael and Cohen-Or, Daniel},
  booktitle=ICLR,
  year={2023},
  url={https://arxiv.org/abs/2208.01626}
}

@inproceedings{tumanyan2023pnp,
  title={Plug-and-Play Diffusion Features for Text-Driven Image-to-Image Translation},
  author={Tumanyan, Narek and Geyer, Michal and Bagon, Shai and Dekel, Tali},
  booktitle=CVPR,
  pages={1921--1930},
  year={2023},
  url={https://openaccess.thecvf.com/content/CVPR2023/html/Tumanyan_Plug-and-Play_Diffusion_Features_for_Text-Driven_Image-to-Image_Translation_CVPR_2023_paper.html}
}

@inproceedings{mokady2023nulltext,
  title={{NULL}-Text Inversion for Editing Real Images Using Guided Diffusion Models},
  author={Mokady, Ron and Hertz, Amir and Aberman, Kfir and Pritch, Yael and Cohen-Or, Daniel},
  booktitle=CVPR,
  pages={6038--6047},
  year={2023},
  url={https://openaccess.thecvf.com/content/CVPR2023/html/Mokady_NULL-Text_Inversion_for_Editing_Real_Images_Using_Guided_Diffusion_Models_CVPR_2023_paper.html}
}

@inproceedings{gerych2024bendvlm,
  title={{BendVLM}: Test-Time Debiasing of Vision-Language Embeddings},
  author={Gerych, Walter and Zhang, Haoran and Hamidieh, Kimia and Pan, Eileen and Sharma, Maanas and Hartvigsen, Thomas and Ghassemi, Marzyeh},
  booktitle=NIPS,
  year={2024},
  url={https://proceedings.neurips.cc/paper_files/paper/2024/hash/72462f451057f9e782af2c3c04c5a4b0-Abstract-Conference.html}
}
}

\clearpage
\setcounter{page}{1}
\maketitlesupplementary
\appendix

\section{Theory}
\label{sec:supp-theory}

We establish the minimum-change interpretation of reward tilting and explain
lookahead weighting and the inherited-score correction in \cref{sec:method}.
Throughout, we fix a prompt $c$ and assume $r(x_0,c)$ is bounded and measurable.

\subsection{Minimum-change reward tilting}
\label{sec:supp-target}

For $\beta\ge0$, the normalized distribution in \cref{eq:target} is
\begin{equation}
\label{eq:supp-normalizer}
\begin{aligned}
p_\beta(x_0\mid c)
&=\frac{p_\theta(x_0\mid c)e^{\beta r(x_0,c)}}{Z_\beta(c)},\\
Z_\beta(c)&=\E_{p_\theta(\cdot\mid c)}[e^{\beta r(x_0,c)}].
\end{aligned}
\end{equation}
Differentiating the log normalizer gives
\begin{equation}
\label{eq:supp-mean-derivative}
\begin{aligned}
\mu_c(\beta)&=\frac{\mathrm d}{\mathrm d\beta}\log Z_\beta(c),\\
\mu_c'(\beta)&=\operatorname{Var}_{p_\beta(\cdot\mid c)}[r(x_0,c)]\ge0.
\end{aligned}
\end{equation}
Thus the mean reward is nondecreasing in the steering strength.

\begin{lemma}[Minimum-change reward tilting]
\label{lem:strength}
If $\mu_c(0)\ge\tau$, then $\beta_c=0$. Otherwise, suppose
$\mu_c(0)<\tau<\operatorname*{ess\,sup}_{p_\theta(\cdot\mid c)}r(x_0,c)$.
Then there is a unique finite $\beta_c>0$ satisfying
$\mu_c(\beta_c)=\tau$. In both cases, $p_{\beta_c}(\cdot\mid c)$
is the unique minimizer in \cref{eq:minimal-change}.
\end{lemma}

\begin{proof}
When $\mu_c(0)\ge\tau$, the base distribution is feasible and uniquely
minimizes KL divergence. Otherwise, finite exponential tilts preserve the base
distribution's null sets, so the variance in \cref{eq:supp-mean-derivative}
is strictly positive. The mean reward increases continuously to the essential
supremum of $r(x_0,c)$ as $\beta\to\infty$. The stated condition thus
yields a unique finite $\beta_c$ with $\mu_c(\beta_c)=\tau$.

For any feasible $Q$ with finite KL divergence, suppressing the conditioning on $c$ for readability,
\begin{equation}
\label{eq:supp-kl-identity}
\begin{aligned}
&\KL(Q\,\|\,p_\theta)-\KL(p_{\beta_c}\,\|\,p_\theta)\\
&\quad=\KL(Q\,\|\,p_{\beta_c})
+\beta_c\big(\E_Q[r]-\mu_c(\beta_c)\big).
\end{aligned}
\end{equation}
The second term is zero when $\beta_c=0$ and nonnegative otherwise, since
$\mu_c(\beta_c)=\tau$ and $Q$ is feasible. The first term is nonnegative,
with equality only when $Q=p_{\beta_c}$. Distributions with infinite KL
divergence cannot improve the objective.
\end{proof}

\subsection{Lookahead weighting of intermediate states}
\label{sec:supp-lookahead-resampling}

The target in \cref{eq:target} concerns completed images, whereas resampling
acts on intermediate denoising states. Consider the base-model distribution
over complete trajectories, reweighted by $e^{\beta r(x_0,c)}$. The weight
for an intermediate state $x_t$ is the expected exponential reward of a full
continuation:
\begin{equation}
\label{eq:supp-ideal-weight}
H_{t,\beta}(x_t,c)
=\E_{x_0\sim p_\theta(\cdot\mid x_t,c)}[e^{\beta r(x_0,c)}].
\end{equation}
Writing $p_\beta(x_t\mid c)$ for the intermediate marginal under this
tilted trajectory law, conditioning on $x_t$ gives
\begin{equation}
\label{eq:supp-ideal-marginal}
p_\beta(x_t\mid c)
=\frac{p_\theta(x_t\mid c)H_{t,\beta}(x_t,c)}{Z_\beta(c)}.
\end{equation}
A preview sampled from the full continuation distribution would make
$e^{\beta r(\tilde x_0,c)}$ an unbiased estimate of $H_{t,\beta}$
conditional on $x_t$. In \methodname{}, the $L$-step coarse preview
instead provides a cheaper proxy; it need not be unbiased for the
full-continuation weight.

\subsection{Inherited scores and adaptive strength}
\label{sec:supp-strength-estimation}

At checkpoint $j$, a selected particle inherits the \emph{log-score}
$h_j=\hat\beta_{c,j}s_j$, with $h_0=0$. The unnormalized incremental factor
in \cref{eq:weights} is $e^{h_j-h_{j-1}}$. Along an ancestral history,
these factors telescope:
\begin{equation}
\label{eq:supp-telescope}
\prod_{j=1}^k e^{h_j-h_{j-1}}=e^{h_k}.
\end{equation}
Thus the incremental factors do not accumulate a separate exponential reward preference from every checkpoint.

To make this precise, fix the checkpoint strengths and consider ideal
population-level resampling. Let $P_k$ denote the base-model law on denoising
histories through checkpoint $k$, including their independently drawn previews.
Immediately before resampling, the population law is proportional to
$e^{h_{k-1}}\,\mathrm dP_k$. At the first checkpoint this holds because
$h_0=0$. Inductively, base-model propagation preserves the inherited
factor, and resampling at checkpoint $j$ multiplies it by
$e^{h_j-h_{j-1}}$. Thus, for any candidate strength $\beta$,
\begin{equation}
\label{eq:supp-preview-mean}
\begin{aligned}
&\frac{\E_{\mathrm{pre},k}[s_k e^{\beta s_k-h_{k-1}}]}{\E_{\mathrm{pre},k}[e^{\beta s_k-h_{k-1}}]}\\
&\qquad=\frac{\E_{P_k}[s_k e^{\beta s_k}]}{\E_{P_k}[e^{\beta s_k}]}.
\end{aligned}
\end{equation}
The right-hand side is the exponentially tilted mean of the base-model
\emph{preview rewards}. It equals $\mu_c(\beta)$ when the base-model
preview rewards have the same distribution as full-continuation rewards; otherwise, it is the
proxy estimated by \cref{eq:level-estimate}.

Finally, for the realized particles, write
$\bar s_k(\beta)=\sum_i\omega_k^i(\beta)s_k^i$. Holding the inherited
scores fixed, differentiating \cref{eq:weights} yields
\begin{equation}
\label{eq:supp-monotone}
\frac{\mathrm d}{\mathrm d\beta}\bar s_k(\beta)
=\sum_{i=1}^n\omega_k^i(\beta)
\big(s_k^i-\bar s_k(\beta)\big)^2\ge0.
\end{equation}
The weighted preview mean is therefore nondecreasing, supporting the
minimum-strength search in \cref{eq:level-estimate} when the threshold
is attainable. These population-level identities explain the weighting rule, but do not
imply exact sampling from $p_{\beta_c}$ with coarse previews, finite
particles, adaptive strengths, or final best-image selection.
\section{Experimental details}
\label{sec:supp-impl}

We describe the shared sampling settings and baseline configurations, followed by the evaluation protocols for text-to-image generation and PIE-Bench. We describe \dsEdit{} separately in \cref{sec:supp-seam-bench}. Code to reproduce our results is provided at \href{https://anonymous.4open.science/r/TrailMix-8045/}{this repository.}

\subsection{Models and sampling}
\label{sec:supp-sampling}
\vspace{5pt}
\noindent
\textbf{Diffusion models.}
We use Stable Diffusion v1.4~\cite{rombach2022ldm} at $512\times512$ resolution and SDXL base 1.0~\cite{podell2024sdxl} at $1024\times1024$. Their classifier-free guidance scales are $7.5$ and $5.0$, respectively. Both models run in half precision with stochastic DDIM sampling~\cite{song2021ddim} ($\eta=1$). Text-to-image generation and generated-source editing use $100$ denoising steps. PIE-Bench editing uses the last $30$ steps of this schedule, initialized with SDEdit~\cite{meng2022sdedit} (\cref{sec:supp-edit-eval}).

\vspace{5pt}
\noindent
\textbf{Seeds and shared initialization.}
For each generation benchmark, we number prompts in dataset order and use the prompt index as the random seed. We draw eight initial noise samples from this seed and share them across methods: Single uses the first, four-particle methods use the first four, and Best-of-8 uses all eight. Methods with different search or pruning procedures use the same initial samples where applicable. We also seed denoising, preview generation, and resampling for reproducibility, but do not synchronize the noise drawn during later denoising steps across methods. Source-image initialization for editing is described in \cref{sec:supp-edit-eval,sec:supp-seam-settings}.

\vspace{5pt}
\noindent
\textbf{\methodname{} previews and resampling.}
For generation, \methodname{} maintains four denoising trajectories and resamples after steps $10$ and $80$. At each checkpoint, it copies each particle's current state and generates one clean-image preview using $L=8$ coarse DDIM steps on evenly spaced timesteps. The reward judge scores these previews, which are then discarded; the original particle states are left unchanged. Preview noise uses a separate random stream from the continuing trajectories.

We then use systematic resampling with the probabilities defined below to select four particles with replacement. Each selected particle resumes from its original checkpoint state, and duplicated particles receive independent denoising noise. We complete all four trajectories after the last checkpoint and return the highest-reward final image.

\vspace{5pt}
\noindent
\textbf{Adaptive steering strength.}
We use ImageReward~\cite{xu2023imagereward} for generation and set the target mean reward to $\tau=1$ for every prompt. At each checkpoint, we estimate the steering strength using only the previews generated at that checkpoint (four in the default configuration). The inherited log-score accounts for previous resampling, but earlier preview rewards are not pooled with the current rewards.

Let $s_k^i=r(\tilde x_{0,k}^i,c)$ be the reward of particle $i$'s preview at checkpoint $k$, and let $h_{k-1}^i$ be the log-score inherited from its parent, initialized to $h_0^i=0$. For a candidate steering strength $\beta\geq0$, the weights and estimated mean reward are
\begin{equation}
\label{eq:supp-current-preview-mean}
\begin{aligned}
\omega_k^i(\beta)
&=\frac{\exp(\beta s_k^i-h_{k-1}^i)}
{\sum_{\ell=1}^n\exp(\beta s_k^\ell-h_{k-1}^\ell)},\\
\hat\mu_{c,k}(\beta)
&=\sum_{i=1}^n\omega_k^i(\beta)s_k^i.
\end{aligned}
\end{equation}
The inherited log-score corrects for selection at earlier checkpoints. Holding these scores fixed, $\hat\mu_{c,k}(\beta)$ is nondecreasing in $\beta$ and approaches the largest preview reward as $\beta\to\infty$.

We choose the smallest nonnegative strength that reaches $\tau$, when such a strength exists. If $\hat\mu_{c,k}(0)\geq\tau$, we set $\hat\beta_{c,k}=0$. Otherwise, if $\max_i s_k^i>\tau$, we find the root $\hat\mu_{c,k}(\hat\beta_{c,k})=\tau$ by bisection. If $\max_i s_k^i\leq\tau$, the target cannot be reached at any finite strength, and we instead use the Gaussian moment approximation
\begin{equation}
\label{eq:supp-strength-fallback}
\hat\beta_{c,k}=\max\!\left\{0,
\frac{\tau-\hat\mu_{c,k}(0)}{\max(\hat\sigma_{c,k}^2,10^{-8})}\right\},
\end{equation}
where $\hat\sigma_{c,k}^2=\sum_i\omega_k^i(0)(s_k^i-\hat\mu_{c,k}(0))^2$. This approximation follows from the mean shift $\beta\hat\sigma_{c,k}^2$ under exponential tilting of Gaussian rewards. It provides a finite steering strength even when the observed preview rewards cannot attain the target. We resample using $\omega_k^i(\hat\beta_{c,k})$, and each selected copy of particle $i$ inherits $h=\hat\beta_{c,k}s_k^i$. The strength is recomputed at the next checkpoint and may increase or decrease.

\vspace{5pt}
\noindent
\textbf{Baseline configurations.}
All reward-based sampling baselines use ImageReward for generation and the editing reward defined in \cref{sec:supp-edit-eval} for editing. Their denoising trajectories use the shared DDIM sampler and initial noises where applicable; LiDAR also uses DPM-Solver for its lookahead samples. We follow the schedules in the original papers or released implementations and use a nominal four-trajectory budget for the sampling and search baselines. Each method allocates this budget according to its selection strategy. We report the actual U-Net and judge-call costs in the result tables.

\emph{Single and Best-of-$N$.}
Single completes one trajectory without reward-based selection. We count one final reward evaluation per output for Single, consistent with the main-paper cost convention. Best-of-$N$ completes $N$ independent trajectories and returns the image with the highest reward. We evaluate Best-of-4 and Best-of-8.

\emph{FK steering}~\cite{singhal2025fk}.
We use the max-potential variant, which tracks the highest reward assigned to a particle's one-step clean-image predictions. Four particles are resampled with steering strength $\lambda=10$ at sampling indices $20$, $40$, $60$, and $80$ (zero-indexed), followed by a terminal scoring checkpoint. These five checkpoints require $20$ judge calls; an additional four calls select the highest-reward completed image. The corresponding inference costs are reported in the main results.

\emph{BFS}~\cite{zhang2026classical}.
We follow the authors' \dsGen{} configuration, which uses the same four particles, max-potential rule, strength, and checkpoints as FK steering. BFS differs in two ways: it uses Srinivasan resampling instead of multinomial resampling, and it multiplies the reward at sampling index $j$ by $2j/T$ before computing the potential. We follow this linear schedule from the released code, rather than the exponential schedule described in the paper. BFS has the same inference costs as FK steering.

\emph{PSP}~\cite{guimaraes2026psp}.
Following the paper's default schedule, PSP begins with eight candidates and prunes them as denoising proceeds. After step $25$, it ranks their one-step clean-image predictions and keeps four candidates; after step $50$, it scores the remaining candidates again and keeps two. It completes both and returns the higher-reward image. The total denoising budget matches Best-of-4 despite the larger initial population.

\emph{SVDD-PM}~\cite{li2024svdd}.
We follow the authors' implementation with $\alpha=0$. At each denoising step except the last, SVDD-PM proposes four next states, scores a one-step clean-image prediction from each, and continues from the highest-reward proposal. It draws one next state at the last step. Unlike four-particle resampling, this procedure maintains one selected path while spending a four-proposal budget at each transition. Its inference costs are reported in the main results.

\emph{DTS}~\cite{jain2025dts}.
We use DTS$^\star$, the greedy variant used for the paper's text-to-image results. It explores a tree of denoising states, either starting from the next initial noise or extending an existing branch. At each node, it selects the child that maximizes its value plus $c\sqrt{\ln N_{\mathrm{parent}}/N_{\mathrm{child}}}$, with $c=0.1$. It then branches by one DDIM step and completes the trajectory. Each finished image is scored once and assigned value $e^r$; internal node values average those of their children. Branching is permitted after $20$, $40$, $60$, and $80\%$ of the steps, with at most $\lfloor2N^{0.4}\rfloor$ children per node of visit count $N$. Search has an $800$-U-Net-evaluation budget, but the last rollout is completed as in the released code, yielding $916$ evaluations and $5$ judge calls. We return the highest-reward completed image.

\emph{LiDAR}~\cite{kim2026lookahead}.
LiDAR first generates and scores $50$ lookahead images $x_0^{(k)}$, with rewards $r_k$, using DPM-Solver from fresh noise. We use five solver steps on SD v1.4 and eight with Karras sigmas on SDXL, as specified in the paper. LiDAR then denoises four particles with the shared DDIM sampler, modifying their predicted noise while $t>200$:
\begin{equation}
\label{eq:supp-lidar}
\begin{aligned}
\hat\epsilon &\leftarrow \hat\epsilon
-s\sqrt{1-\bar\alpha_t}\,\frac{\sqrt{\bar\alpha_t}}{1-\bar\alpha_t}\\
&\qquad\times\sum_{k=1}^{50}\bigl(w_k^{r}-w_k\bigr)x_0^{(k)},\\
w_k&\propto\exp\!\Bigl(-\frac{\lVert x_t-\sqrt{\bar\alpha_t}\,x_0^{(k)}\rVert^2}{2(1-\bar\alpha_t)}\Bigr),\\
w_k^{r}&\propto w_k\,e^{\lambda r_k}.
\end{aligned}
\end{equation}
Both weight vectors are normalized across the $50$ lookaheads. Following the paper's appendix, we use $\lambda=5000$ and correction scale $s=12.5$ on SD v1.4 or $s=8$ on SDXL. We return the highest-reward of the four final images, as in the released code's best-of-$n$ option; the paper's results table instead averages their scores. The costs are $1300$ U-Net evaluations on SD v1.4 and $1600$ on SDXL, with $54$ judge calls for either model.

\vspace{5pt}
\noindent
\textbf{Baseline adaptations for editing.}
For \dsEdit{}, every editing trajectory starts from the initial noise used to generate its source image (\cref{sec:supp-seam-settings}). For PIE-Bench, every trajectory starts from the noised source image (\cref{sec:supp-edit-eval}). The sampling baselines and \methodname{} share these source-conditioned initial states and use independent stochastic noise for subsequent denoising. We replace ImageReward with the editing reward while keeping reward-ranking rules unchanged. No settings for these sampling baselines are tuned on the editing evaluation sets.

We adjust reward scales where necessary. FK steering and BFS use $\lambda=1.21$, calibrated on held-out editing pairs (\cref{sec:supp-seam-settings}). We rescale the LiDAR and DTS$^\star$ reward coefficients by the ratio of the calibrated editing strength to the generation strength (approximately $0.121$). This gives $\lambda=605.5$ for LiDAR and leaf values $e^{0.121\,r}$ for DTS$^\star$. For generated-source edits, these baselines use the full $100$-step trajectory and retain their generation costs.

On PIE-Bench, all methods instead run the final $30$ denoising steps. PSP prunes after steps $8$ and $15$; BFS uses FK steering's editing checkpoints; and DTS$^\star$ branches at $20$, $40$, $60$, and $80\%$ of the shortened trajectory, with a $240$-U-Net-evaluation search budget. For LiDAR, we noise the source to the first sampled timestep and draw source-conditioned lookaheads using five or eight DPM-Solver steps over the remaining range, without Karras sigmas. Its correction is applied at the ten steps with $t>200$. The PIE-Bench costs (U-Net evaluations/judge calls) are $244/14$ for PSP, $240/24$ for BFS, $234/116$ for SVDD-PM, $272/5$ for DTS$^\star$, and $740/54$ for LiDAR on SD v1.4 or $1040/54$ on SDXL.

\subsection{Generation benchmarks and evaluation}
\label{sec:supp-generation-eval}

We evaluate completed images with each benchmark's independent metrics, not with the reward judge used to select images during generation. Scores are reported on a $0$--$100$ scale, with higher values indicating better alignment.

\vspace{5pt}
\noindent
\textbf{RareBench.}
RareBench~\cite{park2025r2f} contains $320$ prompts across eight equally sized categories: five single-object and three multi-object. We follow the R2F rubric, mapping a rating $r$ to $100\max(r-1,0)/4$, using the GPT-4o snapshot \texttt{gpt-4o-2024-11-20}. If the evaluator's content filter blocks an image, we remove that prompt for every method in the comparison. The ImageReward experiments retain $317$ prompts on SD v1.4 and $318$ on SDXL.

\vspace{5pt}
\noindent
\textbf{\dsGen{} and T2I-CompBench++.}
We use the official \dsGen{} object detector and evaluation checks~\cite{ghosh2023geneval} and average the six category scores equally. T2I-CompBench++~\cite{huang2025t2icompbench} contains $1{,}500$ prompts, with $300$ in each of five categories. We report Binding as the mean of the BLIP-VQA scores for color, shape, and texture binding; UniDet evaluates Spatial and Numeracy. When reporting an overall T2I-CompBench++ score, we average all five categories equally, equivalently $(3\,\mathrm{Binding}+\mathrm{Spatial}+\mathrm{Numeracy})/5$.

\subsection{PIE-Bench: editing existing images}
\label{sec:supp-edit-eval}

\vspace{5pt}
\noindent
\textbf{Source initialization.}
PIE-Bench~\cite{ju2024pnp} provides $700$ source images, each with an editing instruction and a mask marking the region to change. We resize each source to the diffusion model's resolution and use SDEdit~\cite{meng2022sdedit} to noise the image before denoising under the target prompt. We use strength $s=0.3$, corresponding to the final $30$ steps of the $100$-step schedule. We selected this strength using Single on SD v1.4 before running the steering comparisons and keep it fixed for both models.

We number source-image/edit pairs in dataset order and use each pair's index as its random seed. All methods start from the same noised source image for a given pair, including Best-of-$N$ and the pruning, search, and guidance baselines. Candidates share this source-conditioned state but receive independent noise during subsequent denoising steps.

\vspace{5pt}
\noindent
\textbf{Editing reward and steering.}
We use Qwen2.5-VL-7B~\cite{bai2025qwen25vl} to ask whether a candidate contains the target attribute and whether it still contains the original attribute. For each question, we convert the probability of ``yes'' to log-odds, clipped to $[-10,10]$. The reward is the target-attribute log-odds minus the original-attribute log-odds, favoring the requested change. For addition-only or deletion-only edits, we use twice the available target log-odds or minus twice the original log-odds, respectively, to retain a comparable reward scale.

\methodname{} uses four particles, eight-step previews, target $\tau=15$, and resampling checkpoints after steps $3$, $14$, and $24$ of the $30$-step trajectory. FK steering uses sampling indices $6$, $12$, $18$, and $24$ (zero-indexed), followed by a terminal checkpoint and a separate final scoring pass, with $\lambda=1.21$ calibrated on $90$ held-out editing pairs (\cref{sec:supp-seam-settings}). The other baselines use the shortened schedules detailed in \cref{sec:supp-sampling}.

\vspace{5pt}
\noindent
\textbf{Edit success.}
We evaluate edits with Gemma~3~27B~\cite{gemma2025gemma3}, independently of the Qwen judge used for steering and final selection. We treat an answer as ``yes'' when its probability is at least $0.5$. A source is valid if it contains the subject and any attribute to be removed, but does not already contain an attribute to be added. On valid sources, an edit succeeds if the output retains the subject and makes the requested addition, deletion, or replacement.

Evaluation questions are available for $698$ examples, of which $514$ have valid sources. EditRate is the percentage of valid sources successfully edited. Context lost is the percentage of valid sources whose outputs no longer contain the subject.

\vspace{5pt}
\noindent
\textbf{Alignment and preservation.}
We compute the official PIE-Bench metrics on outputs resized to $512\times512$. $\mathrm{CLIP}_{\mathrm{edit}}$ evaluates target-text alignment within the edit mask, Structure measures changes in DINO-ViT self-similarity~\cite{tumanyan2022splicing}, and PSNR and LPIPS assess preservation outside the edit mask where an unedited region exists. Higher $\mathrm{CLIP}_{\mathrm{edit}}$ and PSNR, and lower Structure and LPIPS, are better. Following \cref{tab:edit}, Structure and LPIPS are multiplied by $10^3$ in all tables.

\subsection{Inference cost}
\label{sec:supp-cost}

We report denoiser evaluations and reward-judge calls separately. With classifier-free guidance, each DDIM step requires two U-Net evaluations. For $n$ particles, $T$ denoising steps, $K$ checkpoints, and one $L$-step preview per particle at each checkpoint, \methodname{} requires
\begin{equation}
N_{\mathrm{U\text{-}Net}}=2nT+2nKL,
\qquad N_{\mathrm{judge}}=n(K+1).
\end{equation}
These counts include all previews and the final image-selection pass: $928$ U-Net evaluations and $12$ judge calls per generation prompt, $992$ and $16$ per \dsEdit{} edit, and $432$ and $16$ per PIE-Bench edit.

\section{SEAM-Bench Benchmark}
\label{sec:supp-seam-bench}
\label{sec:supp-seam}

\dsEdit{} (Semantic Editing and Attribute Maintenance Benchmark) evaluates whether a requested attribute change succeeds while unrelated image details are preserved. Here, we describe the benchmark construction, source-conditioned editing setup, and evaluation protocol.

\subsection{Dataset construction}
\label{sec:supp-seam-data}
\begin{figure*}[t]
\centering
\includegraphics[width=0.85\textwidth]{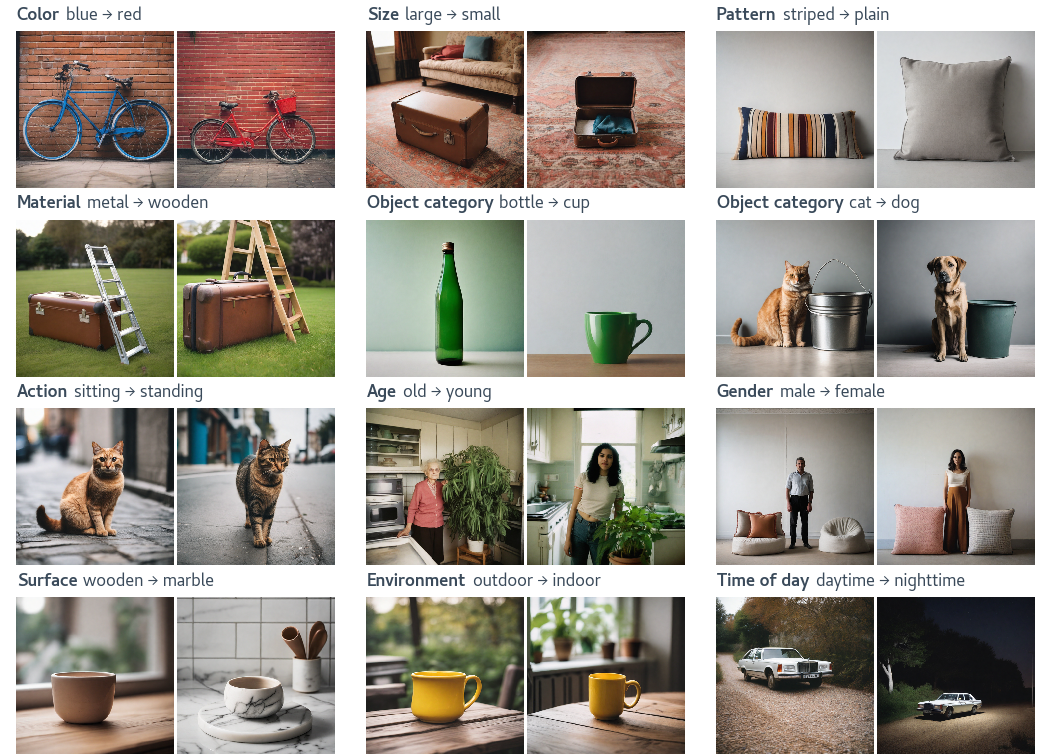}
\caption{\textbf{Examples of the $12$ edit types in \dsEdit{}.} Each pair shows the source (left) and edited image (right), labeled with the requested change. These are selected SDXL outputs from \methodname{} that Gemma~3 judges successful (\cref{sec:supp-seam-metrics}). The examples illustrate edit success; other image details may also change.}
\label{fig:supp-seam-examples}
\end{figure*}

\vspace{5pt}
\noindent
\textbf{Controlled description pairs.}
Each edit pairs source and target descriptions that differ in one semantic attribute while specifying the same remaining scene content. For instance, the color edit in \cref{fig:supp-seam-examples} changes a blue bicycle to a red bicycle, with the bicycle and its surroundings intended to remain unchanged. The benchmark covers $12$ edit types, each defined by two attribute values (\cref{tab:supp-seam-types}). A \emph{context} specifies the shared scene details, and initially defines two \emph{directed edits}, one for each direction of the attribute change.

For each edit type, we divide contexts into disjoint fitting, calibration, and evaluation sets. We use the calibration set to determine FK steering's editing strength (\cref{sec:supp-seam-settings}). The fitting contexts remain excluded from evaluation. All reported results use the evaluation set.

We initially select $16$ evaluation contexts per edit type and include both directions, yielding $384$ directed edits. Removing duplicate source--target pairs leaves $356$ edits. We further exclude overlaps with the fitting and calibration sets, including target-description matches, leaving $316$ edits across $159$ contexts: $157$ with both directions and two with only one. \cref{tab:supp-seam-types} gives the counts by edit type, and \cref{fig:supp-seam-examples} illustrates the requested changes across all $12$ types.

\begin{table}[h]
\centering
\small
\caption{\textbf{\dsEdit{} edit types.} Each type switches between two attribute values; edits are directed, and most contexts are edited in both directions.}
\label{tab:supp-seam-types}
\begin{tabular}{llr}
\toprule
Edit type & Attribute values & Directed edits \\
\midrule
Color & red, blue & 32 \\
Size & small, large & 32 \\
Pattern & striped, plain & 32 \\
Material & wooden, metal & 22 \\
Object category & cup, bottle & 32 \\
Object category & dog, cat & 14 \\
Action & sitting, standing & 30 \\
Age & younger, older & 24 \\
Gender & male, female & 12 \\
Surface & wooden, marble & 24 \\
Environment & indoor, outdoor & 30 \\
Time of day & daytime, nighttime & 32 \\
\midrule
Total & & 316 \\
\bottomrule
\end{tabular}
\end{table}

\vspace{5pt}
\noindent
\textbf{Evaluation questions.}
For each edit, we construct three types of yes/no questions from the dataset record before examining model outputs. These questions are shared across methods and distinguish whether the requested change succeeds from whether other scene details are preserved. We illustrate each question type using the bicycle color edit in \cref{fig:supp-seam-examples}; the questions below are illustrative.
\begin{itemize}
\item \textbf{Subject:} Is there a bicycle in the image? The subject is the object or person carrying the edited attribute. For occupational edits, we identify the person by their occupation; for object-category edits, we use a shared category (e.g., ``animal''); and for scene-level edits, we use the main object.
\item \textbf{Original and target attributes:} Is the bicycle blue? Is the bicycle red? These questions assess whether the original attribute is removed and the target attribute is introduced, respectively.
\item \textbf{Preservation:} Is there a brick wall behind the bicycle? Questions of this type assess scene details unrelated to the requested change.
\end{itemize}
To isolate preservation from edit success, we exclude preservation questions that concern the edited attribute, mention either of its values, or duplicate the subject question. We replace questions requiring direct comparison with a reference image with four single-image questions about lighting, framing, background, and color. Each edit retains $7$ to $13$ preservation questions. The same questions are applied independently to the source and edited images, as described in \cref{sec:supp-seam-metrics}.

\vspace{5pt}
\noindent
\textbf{Counter-stereotypical extension.}
We also construct a separate cohort of $192$ edits that move toward or away from stereotypical associations. It covers six association types: occupation--gender (``doctor''), occupation--age (``judge''), activity--gender (``laying bricks''), and three object associations involving pattern, material, or color (a striped deck chair, a wooden broom handle, and a red fire extinguisher). Each record specifies the subject, its stereotypical attribute value, and a physically plausible alternative; these assignments are fixed before examining model outputs. A \emph{toward} edit introduces the stereotypical value, while an \emph{against} edit replaces it with the alternative.

Each type contains $30$ contexts, of which $16$ are selected for evaluation using the same rule as the main benchmark. Evaluating both directions yields $96$ toward and $96$ against edits. Both directions use the same source-validity checks and success criteria.

\subsection{Experimental settings}
\label{sec:supp-seam-settings}

\vspace{5pt}
\noindent
\textbf{Source generation and initialization.}
We generate source images from the source descriptions separately for SD v1.4 and SDXL, using the settings in \cref{sec:supp-sampling} and $100$ denoising steps. For every edit, the source's initial noise is generated with random seed $0$.

For each edit, all sampling methods initialize every trajectory with the same initial noise used to generate the source image and denoise under the target description. The trajectories differ in the stochastic noise drawn during subsequent denoising, not in their initial states. We evaluate preservation of the source separately from edit success (\cref{sec:supp-seam-metrics}).

\vspace{5pt}
\noindent
\textbf{Editing reward.}
We use the Qwen2.5-VL editing reward defined in \cref{sec:supp-edit-eval}, which favors introducing the target attribute and removing the original attribute.

\vspace{5pt}
\noindent
\textbf{Sampling configurations.}
The sampling baselines use the full-trajectory editing configurations in \cref{sec:supp-sampling}.

To account for the editing reward's different scale, we calibrate FK steering's strength on $90$ held-out edits. At the first intermediate checkpoint, both ImageReward and the editing judge score the four particles' one-step clean-image predictions. For each judge, we compute the difference between the highest and lowest particle rewards, then take the median of these differences across calibration edits. We then multiply FK steering's ImageReward strength, $\lambda=10$, by the ratio of the median ImageReward spread to the median editing-reward spread. This yields $\lambda=1.21$ for editing. None of the calibration edits is included in evaluation.

\methodname{} uses four trajectories, with resampling checkpoints after steps $10$, $45$, and $80$. It generates one eight-step preview per particle at each checkpoint and estimates the steering strength from those preview rewards as described in \cref{sec:supp-sampling}, using target $\tau=15$. After denoising, we return the highest-reward image. We use the same configuration for counter-stereotypical edits.

\vspace{5pt}
\noindent
\textbf{Inference cost.}
We report U-Net evaluations and judge calls as defined in \cref{sec:supp-cost}. Source-image generation is shared across methods and excluded from editing costs.

\subsection{Evaluation protocol}
\label{sec:supp-seam-metrics}

\vspace{5pt}
\noindent
\textbf{Source validity and edit success.}
We evaluate edited images using Gemma~3~27B~\cite{gemma2025gemma3}, independently of the Qwen judge used during sampling and final selection. An answer counts as ``yes'' when its probability is at least $0.5$.

We retain only sources that contain the subject and original attribute but not the target attribute. This ensures that the requested edit requires a change. This leaves $240$ valid sources for SD v1.4 and $283$ for SDXL.

An edit succeeds if the output retains the subject, removes the original attribute, and introduces the target attribute. EditRate is the percentage of valid sources successfully edited; Context lost is the percentage of valid sources whose edited outputs no longer contain the subject.

\vspace{5pt}
\noindent
\textbf{Preservation of unchanged attributes.}
For each preservation question, we compare Gemma's probability of answering ``yes'' on the source and edited images and take the absolute difference. We average these differences within each edit, then average across edits with valid sources. This gives every edit equal weight, regardless of how many preservation questions it contains. The resulting preservation error $E$ ranges from $0$ to $1$; lower values indicate better preservation. We also report $E_{\mathrm{verified}}$, calculated in the same way but restricted to successful edits. We report both preservation metrics on their original $0$--$1$ scale throughout.

\vspace{5pt}
\noindent
\textbf{Directional asymmetry.}
We check source validity separately for the two editing directions. On SD v1.4, $66$ sources are valid for toward-stereotype edits and $80$ for against-stereotype edits; the corresponding counts for SDXL are $74$ and $90$. Each directional EditRate uses its own valid-source set. To measure asymmetry on comparable examples, we instead restrict both directions to contexts with valid sources in each direction: $57$ contexts for SD v1.4 and $69$ for SDXL. Asymmetry is the absolute difference between the two EditRates computed on this shared set, rather than on separately validated sources. The corresponding signed difference is reported in \cref{sec:supp-bias}.

\section{Analysis and ablations}
\label{sec:supp-ablations}

We first investigate the two mechanisms underlying \methodname{}: lookahead-based reward estimation and adaptive steering strength. We then characterize its sensitivity to sampling configurations and assess whether improvements extend to alternative reward judges and evaluators. Finally, we examine how these design choices affect targeted image editing, where edit success must be considered alongside source preservation. Unless stated otherwise, we use SD v1.4 and the task-specific settings in \cref{sec:supp-sampling,sec:supp-generation-eval,sec:supp-edit-eval,sec:supp-seam-settings,sec:supp-seam-metrics}, varying only the stated configurations. Generation scores follow the main-paper evaluation protocol; Avg. weights the three benchmarks equally, and T2I denotes T2I-CompBench++. Bold indicates the best metric within each comparable block.

\subsection{Lookahead-based reward estimation}
\label{sec:supp-rank}
\label{sec:st-preview}

Lookahead previews aim to reveal semantic details that are not yet clear in one-step predictions. We investigate whether they improve the identification of promising trajectories, how preview length affects final alignment, and how best to allocate preview computation.

\vspace{5pt}
\noindent
\textbf{Lookaheads improve early trajectory ranking.}
To isolate reward estimation from resampling, we generate eight independent trajectories per prompt without resampling, using prompt-index seeds offset by $3{,}000{,}000$. At the steps listed in \cref{tab:supp-rank}, we rank the same trajectories using either a one-step prediction or an eight-step preview. We complete the trajectories and determine their final success using independent benchmark evaluators. These analysis checkpoints do not change the sampling process.

We measure the fraction of eventual successes ranked among the top four of eight, restricting the analysis to prompts with both successful and unsuccessful outputs. We pool successful trajectories across these prompts and assign fractional credit to ties at the cutoff. Success means passing the official \dsGen{} checks, obtaining a RareBench rating of at least $4/5$, or satisfying the relevant T2I-CompBench++ criterion (BLIP-VQA binding $\geq0.5$, UniDet spatial $>0$, or numeracy $=1$). These binary criteria are used for the trajectory analysis, not the main benchmark scores.

Across all three benchmarks, eight-step previews identify more eventual successes than one-step predictions early in denoising, while their rankings become similar near completion (\cref{tab:supp-rank}). This demonstrates the value of lookahead evaluation as a selection signal independently of resampling.

\begin{table}[!ht]
\centering
\scriptsize
\setlength{\tabcolsep}{2.5pt}
\begin{tabular}{@{}lcccccc@{}}
\toprule
Scoring method & $10$ & $20$ & $30$ & $40$ & $60$ & $80$\\
\midrule
\multicolumn{7}{@{}l}{\emph{\dsGen: 243 mixed-outcome queries}}\\
One-step & 53.7 & 60.0 & 63.8 & 65.4 & 67.6 & \textbf{67.9}\\
Preview ($L=8$) & \textbf{62.2} & \textbf{63.7} & \textbf{65.6} & \textbf{67.0} & \textbf{67.7} & 67.8\\
\midrule
\multicolumn{7}{@{}l}{\emph{RareBench: 182 mixed-outcome queries}}\\
One-step & 51.6 & 54.1 & 58.9 & 62.3 & 64.0 & \textbf{66.7}\\
Preview ($L=8$) & \textbf{54.9} & \textbf{60.7} & \textbf{62.7} & \textbf{64.5} & \textbf{66.4} & 66.3\\
\midrule
\multicolumn{7}{@{}l}{\emph{T2I-CompBench++: 799 mixed-outcome queries}}\\
One-step & 52.9 & 59.8 & 63.6 & 65.4 & 67.9 & 68.2\\
Preview ($L=8$) & \textbf{59.6} & \textbf{63.2} & \textbf{65.7} & \textbf{66.4} & \textbf{68.6} & \textbf{68.6}\\
\bottomrule
\end{tabular}
\caption{\textbf{Ranking eventual successes.} Percentage of successful trajectories ranked top four of eight, on mixed-outcome prompts. Columns give the denoising step. No resampling; chance level is $50\%$.}
\label{tab:supp-rank}
\end{table}

\begin{table}[!ht]
\centering
\scriptsize
\setlength{\tabcolsep}{2.5pt}
\begin{tabular}{@{}lccccc@{}}
\toprule
$L$ & \dsGen & RareBench & T2I & Avg. & U-Net/judge\\
\midrule
$1$ & 54.4 & 57.0 & 44.8 & 52.1 & 816/12\\
$2$ & 53.5 & 56.7 & 45.4 & 51.9 & 832/12\\
$4$ & 54.8 & 61.1 & 47.2 & 54.4 & 864/12\\
$8$ & \textbf{56.3} & \textbf{61.4} & \textbf{50.7} & \textbf{56.1} & 928/12\\
$16$ & 55.6 & 58.9 & 45.7 & 53.4 & 1056/12\\
\bottomrule
\end{tabular}
\caption{\textbf{Preview length in generation.} Four particles, $\tau=1$. T2I denotes T2I-CompBench++; Avg. weights the three benchmarks equally.}
\label{tab:supp-preview-length}
\end{table}

\vspace{5pt}
\noindent
\textbf{Short lookaheads improve alignment without additional judge calls.}
Varying the preview length, we find that eight-step previews achieve the strongest alignment across all three generation benchmarks. Extending previews to sixteen steps increases denoising cost without improving alignment (\cref{tab:supp-preview-length}).

\vspace{5pt}
\noindent
\textbf{Longer previews outperform multiple shorter previews at matched denoising cost on GenEval.}
\label{sec:supp-grid}
\label{sec:st-grid}
We compare longer previews with multiple independent continuations per particle. For multiple previews, we average their exponential reward scores before resampling, using only previews from the current checkpoint. At matched denoising cost, a single eight-step preview achieves higher \dsGen{} alignment than the tested combinations of shorter previews and requires fewer judge calls (\cref{tab:supp-preview-budget}). Generating four eight-step previews offers only a small further gain at higher cost.

\begin{table}[!ht]
\centering
\scriptsize
\setlength{\tabcolsep}{5pt}
\begin{tabular}{@{}cccc@{}}
\toprule
$m$ & $L$ & \dsGen$\uparrow$ & U-Net/judge\\
\midrule
1 & 2 & 53.5 & 832/12\\
1 & 4 & 54.8 & 864/12\\
1 & 8 & 56.3 & 928/12\\
2 & 2 & 54.0 & 864/20\\
2 & 4 & 53.3 & 928/20\\
2 & 8 & 55.5 & 1056/20\\
4 & 2 & 54.6 & 928/36\\
4 & 4 & 54.1 & 1056/36\\
4 & 8 & \textbf{56.6} & 1312/36\\
\bottomrule
\end{tabular}
\caption{\textbf{Allocation of preview computation.} Four particles; $m$ previews per particle at each checkpoint.}
\label{tab:supp-preview-budget}
\end{table}

\subsection{Adaptive steering strength}
\label{sec:supp-strength}
\label{sec:st-strength}

We examine whether adapting steering strength improves alignment at a fixed preview and resampling budget.

\vspace{5pt}
\noindent
\textbf{Adaptive steering improves aggregate alignment while retaining more lineages than strong fixed steering.}
We compare adaptive and fixed strengths at identical preview and resampling budgets. We count surviving ancestors as the number of distinct initial trajectories represented in the final particle population, a measure of lineage retention rather than visual diversity. Adaptive steering matches or exceeds the best tested fixed setting on each benchmark (\cref{tab:supp-strength}). Weak fixed steering retains more lineages than adaptive steering but achieves lower aggregate alignment.

\begin{table}[!ht]
\centering
\scriptsize
\setlength{\tabcolsep}{2.5pt}
\begin{tabular}{@{}lccccc@{}}
\toprule
Steering strength & \dsGen & RareBench & T2I & Avg. & Ancestors\\
\midrule
Adaptive & \textbf{56.3} & \textbf{61.4} & \textbf{50.7} & \textbf{56.1} & 1.95\\
Fixed $\beta=1$ & 55.2 & \textbf{61.4} & 45.6 & 54.1 & \textbf{2.59}\\
Fixed $\beta=2.5$ & 55.7 & 60.7 & 46.5 & 54.3 & 1.82\\
Fixed $\beta=5$ & 54.9 & 60.5 & 47.1 & 54.2 & 1.41\\
Fixed $\beta=10$ & 55.7 & 59.2 & 47.1 & 54.0 & 1.17\\
Fixed $\beta=20$ & \textbf{56.3} & 60.0 & 47.3 & 54.5 & 1.06\\
\bottomrule
\end{tabular}
\caption{\textbf{Adaptive versus fixed steering.} SD v1.4, four particles, $L=8$; all settings use $928$ U-Net evaluations and $12$ judge calls. T2I denotes T2I-CompBench++; ancestors count surviving initial lineages.}
\label{tab:supp-strength}
\end{table}

\subsection{Sensitivity to sampling configurations}
\label{sec:supp-ablation}
\label{sec:st-tau}

We characterize sensitivity to the target mean reward, resampling schedule, and sampling budgets.

\vspace{5pt}
\noindent
\textbf{A higher target reward does not necessarily improve alignment.}
Increasing $\tau$ reduces independently evaluated alignment across all three benchmarks with eight-step previews, while shorter previews yield mixed changes (\cref{tab:supp-threshold}).

\begin{table}[!ht]
\centering
\scriptsize
\setlength{\tabcolsep}{3pt}
\begin{tabular}{@{}ccccc@{}}
\toprule
$L$ & $\tau$ & \dsGen & RareBench & T2I\\
\midrule
8 & 1 & \textbf{56.3} & \textbf{61.4} & \textbf{50.7}\\
8 & 1.5 & 55.7 & 61.2 & 46.1\\
\midrule
2 & 1 & 53.5 & 56.7 & \textbf{45.4}\\
2 & 1.5 & \textbf{54.4} & \textbf{57.3} & 45.1\\
\bottomrule
\end{tabular}
\caption{\textbf{Sensitivity to the target mean reward.} Four particles; inference costs are identical within each preview length. Bold marks the best score within each block.}
\label{tab:supp-threshold}
\end{table}

\vspace{5pt}
\noindent
\textbf{Alignment varies modestly across the tested resampling schedules.}
\label{sec:supp-resampling-schedule}
We vary the number and placement of resampling checkpoints while keeping the other generation settings fixed. At each fixed checkpoint count $K$, schedules that begin resampling earlier achieve stronger alignment (\cref{tab:supp-resampling-schedule}). The best three- and four-checkpoint schedules provide only small gains over the default at higher denoising and judge cost, supporting the two-checkpoint configuration as a favorable tradeoff.

\begin{table}[!ht]
\centering
\scriptsize
\setlength{\tabcolsep}{2.5pt}
\begin{tabular}{@{}lcccc@{}}
\toprule
Checkpoints & \dsGen & RareBench & T2I & U-Net/judge\\
\midrule
\multicolumn{5}{@{}l}{\emph{$K=2$}}\\
10, 80 (default) & \textbf{56.3} & \textbf{61.4} & \textbf{50.7} & 928/12\\
20, 80 & 56.0 & 61.0 & 50.5 & 928/12\\
20, 60 & 55.8 & 60.6 & 50.2 & 928/12\\
40, 80 & 55.1 & 59.8 & 49.6 & 928/12\\
\midrule
\multicolumn{5}{@{}l}{\emph{$K=3$}}\\
10, 45, 80 & \textbf{56.6} & \textbf{61.8} & \textbf{50.9} & 992/16\\
20, 50, 80 & 56.2 & 61.2 & 50.6 & 992/16\\
\midrule
\multicolumn{5}{@{}l}{\emph{$K=4$}}\\
10, 30, 55, 80 & \textbf{56.7} & \textbf{61.9} & \textbf{51.0} & 1056/20\\
20, 40, 60, 80 & 56.4 & 61.5 & 50.8 & 1056/20\\
\bottomrule
\end{tabular}
\caption{\textbf{Sensitivity to the resampling schedule.} SD v1.4, four particles, $L=8$. Checkpoints are denoising step counts; $K$ denotes their number. Bold marks the best score within each checkpoint count.}
\label{tab:supp-resampling-schedule}
\end{table}

\vspace{5pt}
\noindent
\textbf{Steering is more effective with four or eight particles than with two.}
\label{sec:supp-particles}
\label{sec:st-particles}
\label{sec:supp-steps}
\label{sec:st-steps}
\methodname{} outperforms the corresponding FK steering and Best-of-$N$ baselines on both compositional benchmarks with four and eight particles, but the results are mixed with two particles (\cref{tab:supp-particles}). With two particles, both final particles share an initial ancestor for $87\%$ of prompts, suggesting that very small populations leave less room for selection. This complements the larger-population scaling analysis in the main paper.

\textbf{More denoising steps do not consistently improve alignment.} We vary the trajectory length while preserving each method's relative checkpoint positions. \methodname{} exceeds FK steering on RareBench at every tested denoising length, while its advantage over Best-of-4 depends on the budget (\cref{tab:supp-particles}).

\begin{table}[!ht]
\centering
\scriptsize
\begin{tabular}{@{}lcccc@{}}
\toprule
Method & \dsGen & RareBench & T2I & U-Net/judge\\
\midrule
\multicolumn{5}{@{}l}{\emph{$n=2$, $T=100$}}\\
Best-of-2 & \textbf{50.6} & \textbf{53.6} & 40.5 & 400/2\\
FK steering & 50.1 & 52.1 & 39.8 & 400/12\\
\methodname{} & 48.6 & 53.5 & \textbf{41.7} & 464/6\\
\midrule
\multicolumn{5}{@{}l}{\emph{$n=4$, $T=100$}}\\
Best-of-4 & 53.9 & 56.3 & 46.1 & 800/4\\
FK steering & 53.0 & 54.5 & 44.7 & 800/24\\
\methodname{} & \textbf{56.3} & \textbf{61.4} & \textbf{50.7} & 928/12\\
\midrule
\multicolumn{5}{@{}l}{\emph{$n=8$, $T=100$: compositional benchmarks}}\\
Method & \dsGen & \multicolumn{2}{c}{T2I} & U-Net/judge\\
Best-of-8 & 56.3 & \multicolumn{2}{c}{50.5} & 1600/8\\
FK steering & 56.8 & \multicolumn{2}{c}{48.8} & 1600/48\\
\methodname{} & \textbf{58.0} & \multicolumn{2}{c}{\textbf{51.2}} & 1856/24\\
\midrule
\multicolumn{5}{@{}l}{\emph{$n=4$, $T=50$}}\\
Method & \dsGen & RareBench & T2I & U-Net/judge\\
Best-of-4 & \textbf{54.9} & 59.2 & \textbf{45.9} & 400/4\\
FK steering & 54.7 & 55.6 & 44.9 & 400/24\\
\methodname{} & 54.6 & \textbf{59.7} & 45.8 & 528/12\\
\midrule
\multicolumn{5}{@{}l}{\emph{$n=4$, $T=200$}}\\
Best-of-4 & 52.0 & \textbf{61.3} & 45.7 & 1600/4\\
FK steering & 54.6 & 57.2 & 44.7 & 1600/24\\
\methodname{} & \textbf{54.8} & 59.9 & \textbf{47.1} & 1728/12\\
\bottomrule
\end{tabular}
\caption{\textbf{Particle and denoising budgets.} SD v1.4, $L=8$. The eight-particle block reports compositional benchmarks. RareBench coverage differs slightly at $T=50$. Bold indicates the best reported score within each budget. T2I denotes T2I-CompBench++.}
\label{tab:supp-particles}
\end{table}

\subsection{Robustness to reward and evaluation models}
\label{sec:supp-vlm}
\label{sec:st-judge}
\label{sec:supp-gemma}
\label{sec:st-evaluator}

We test whether the improvements observed with ImageReward and the original benchmark evaluators persist under alternative scoring models. We first change the reward used during generation and then evaluate the resulting images using an independent model.

\vspace{5pt}
\noindent
\textbf{The gains persist across alternative reward judges on all three benchmarks.}
We replace ImageReward with Qwen2.5-VL-7B~\cite{bai2025qwen25vl} or PickScore~\cite{kirstain2023pickapic} on SD v1.4, and also test Qwen2.5-VL on SDXL. Within each comparison, methods use the same reward judge and independent benchmark evaluation; \methodname{} uses four particles and eight-step previews. We use judge-specific configurations as follows.

For Qwen2.5-VL, we ask ``Does this image show $\langle\text{query}\rangle$?'' and clip the log-odds of ``yes'' to $[-10,10]$. We use $\tau=7.5$ and checkpoints $10$, $45$, and $80$; FK steering uses $\lambda=1.21$ and its standard five-checkpoint schedule. RareBench retains $317$ prompts on SD v1.4 and $315$ on SDXL after content-filter exclusions.

For PickScore, we map the raw score $\mathrm{PS}$ to ImageReward's scale via $0.7324\,\mathrm{PS}-14.715$, allowing the ImageReward settings ($\tau=1$, checkpoints $10$ and $80$, and FK $\lambda=10$) to carry over. The mapping uses $400$ SD v1.4 Best-of-4 images from the $100$ ImageReward-benchmark prompts, outside our evaluation sets; PickScore and ImageReward means (standard deviations) on these images are $20.42$ ($1.23$) and $0.241$ ($0.902$), respectively.

\methodname{} exceeds both four-particle baselines (Best-of-4 and FK steering) on GenEval, RareBench, and T2I-CompBench++ for every tested alternative reward judge and backbone (\cref{tab:supp-judges}). Best-of-8 remains stronger on several metrics, but uses twice as many full denoising trajectories.

\begin{table}[!ht]
\centering
\scriptsize
\begin{tabular}{@{}lcccc@{}}
\toprule
Method & \dsGen & RareBench & T2I & U-Net/judge\\
\midrule
\multicolumn{5}{@{}l}{\emph{SD v1.4: Qwen2.5-VL}}\\
Best-of-4 & 55.9 & 58.7 & 47.1 & 800/4\\
Best-of-8 & \textbf{60.6} & \textbf{64.3} & \textbf{51.7} & 1600/8\\
FK steering & 55.1 & 57.3 & 46.9 & 800/24\\
\methodname{} & 57.7 & 64.1 & 48.2 & 992/16\\
\midrule
\multicolumn{5}{@{}l}{\emph{SD v1.4: PickScore}}\\
Best-of-4 & 52.9 & 55.6 & 42.3 & 800/4\\
Best-of-8 & \textbf{56.9} & \textbf{58.0} & \textbf{45.0} & 1600/8\\
FK steering & 50.3 & 53.3 & 41.9 & 800/24\\
\methodname{} & 53.2 & \textbf{58.0} & 42.7 & 928/12\\
\midrule
\multicolumn{5}{@{}l}{\emph{SDXL: Qwen2.5-VL}}\\
Best-of-4 & 68.3 & 74.6 & 56.6 & 800/4\\
Best-of-8 & \textbf{73.5} & 76.9 & \textbf{59.9} & 1600/8\\
FK steering & 68.3 & 72.5 & 56.5 & 800/24\\
\methodname{} & 68.8 & \textbf{77.5} & 57.4 & 992/16\\
\bottomrule
\end{tabular}
\caption{\textbf{Alternative reward judges.} \methodname{} uses three checkpoints with Qwen2.5-VL and two with PickScore. Scores use independent evaluators; bold marks the best score within each model--judge block. T2I denotes T2I-CompBench++.}
\label{tab:supp-judges}
\vspace{-10pt}
\end{table}

\vspace{5pt}
\noindent
\textbf{Independent evaluation corroborates gains across all three benchmarks.}
We rescore the same returned images from ImageReward- and Qwen-guided runs using Gemma~3~27B~\cite{gemma2025gemma3}. We ask ``Does this image show $\langle\text{query}\rangle$?'' and report the mean probability of ``yes'' multiplied by $100$. Under Gemma~3 evaluation, \methodname{} exceeds both four-particle baselines on GenEval, RareBench, and T2I-CompBench++ across both backbones and both reward judges (\cref{tab:supp-gemma}). Best-of-8 remains stronger on several metrics at its larger sampling budget. These results indicate that the improvements over the four-particle baselines are not specific to the original evaluators.

\begin{table}[!ht]
\centering
\scriptsize
\setlength{\tabcolsep}{3pt}
\begin{tabular}{@{}lccc@{}}
\toprule
Method & \dsGen & RareBench & T2I\\
\midrule
\multicolumn{4}{@{}l}{\emph{SD v1.4: ImageReward}}\\
Best-of-4 & 69.3 & 19.8 & 43.6\\
Best-of-8 & \textbf{72.4} & 24.6 & \textbf{49.1}\\
FK steering & 66.7 & 20.2 & 43.4\\
\methodname{} & 69.9 & \textbf{25.5} & 45.9\\
\midrule
\multicolumn{4}{@{}l}{\emph{SD v1.4: Qwen2.5-VL}}\\
Best-of-4 & 72.4 & 22.4 & 51.1\\
Best-of-8 & \textbf{78.2} & \textbf{29.9} & \textbf{56.6}\\
FK steering & 69.6 & 25.5 & 49.1\\
\methodname{} & 72.6 & 29.3 & 53.8\\
\midrule
\multicolumn{4}{@{}l}{\emph{SDXL: ImageReward}}\\
Best-of-4 & 79.9 & 33.6 & 56.6\\
Best-of-8 & \textbf{83.0} & 37.2 & \textbf{60.5}\\
FK steering & 80.5 & 36.8 & 56.0\\
\methodname{} & 82.7 & \textbf{37.7} & 57.4\\
\midrule
\multicolumn{4}{@{}l}{\emph{SDXL: Qwen2.5-VL}}\\
Best-of-4 & 83.7 & 41.8 & 65.6\\
Best-of-8 & \textbf{89.4} & 46.0 & \textbf{72.5}\\
FK steering & 82.9 & 41.8 & 63.5\\
\methodname{} & 86.3 & \textbf{48.4} & 67.6\\
\bottomrule
\end{tabular}
\caption{\textbf{Gemma~3 independent evaluation.} Mean probability of a positive full-prompt judgment, multiplied by $100$. T2I denotes T2I-CompBench++; bold marks the best score per column.}
\label{tab:supp-gemma}
\vspace{-10pt}
\end{table}

\subsection{Analysis of targeted image editing}
\label{sec:supp-edit-analysis}

We examine whether the design choices studied in generation have similar effects on editing, and how steering relates to the tradeoff between edit success and source preservation.

\vspace{5pt}
\noindent
\textbf{Edit success varies modestly across preview configurations.}
On \dsEdit{}, longer or more numerous previews yield limited gains in EditRate (\cref{tab:supp-preview-edit}). This is consistent with the smaller early ranking gain relative to generation when we repeat the no-resampling analysis with eight source-conditioned trajectories per edit, using independent subsequent denoising noise. Qwen2.5-VL scores the previews and Gemma~3 determines success. Among the $104$ mixed-outcome edits, eight-step previews rank $57.4\%$ of eventual successes among the top four at step $10$, compared with $55.7\%$ for one-step predictions. On PIE-Bench, longer previews modestly improve edit success with comparable preservation.

\textbf{Adaptive steering achieves competitive editing success across the tested strengths.} With two-step previews and equal inference budgets, adaptation achieves the highest \dsEdit{} EditRate among the tested steering settings, while the best-performing fixed settings achieve similar success and the weakest fixed strength yields the lowest preservation error. Adaptation therefore achieves competitive editing performance without selecting a fixed strength from the grid. Increasing the target reward does not improve \dsEdit{} EditRate and produces only small changes on PIE-Bench (\cref{tab:supp-preview-edit}).

\begin{table}[!ht]
\centering
\scriptsize
\setlength{\tabcolsep}{4pt}
\begin{tabular}{@{}lccc@{}}
\toprule
Configuration & EditRate$\uparrow$ & $E\downarrow$ & U-Net/judge\\
\midrule
\multicolumn{4}{@{}l}{\emph{\dsEdit{}: preview count and length $(m,L)$}}\\
$(1,1)$ & 90.4 & 0.221 & 824/16\\
$(1,2)$ & 92.5 & 0.222 & 848/16\\
$(1,4)$ & 91.2 & \textbf{0.212} & 896/16\\
$(1,8)$ & 92.5 & 0.217 & 992/16\\
$(2,2)$ & 92.5 & 0.226 & 896/28\\
$(2,4)$ & 92.1 & 0.220 & 992/28\\
$(2,8)$ & 93.3 & 0.220 & 1184/28\\
$(4,2)$ & 93.3 & 0.233 & 992/52\\
$(4,4)$ & 92.9 & 0.221 & 1184/52\\
$(4,8)$ & \textbf{93.8} & 0.216 & 1568/52\\
\midrule
\multicolumn{4}{@{}l}{\emph{\dsEdit{}: steering strength ($L=2$)}}\\
Adaptive, $\tau=15$ & \textbf{92.5} & 0.222 & 848/16\\
Adaptive, $\tau=18$ & 91.7 & 0.228 & 848/16\\
Fixed, $\beta=0.5$ & 91.2 & \textbf{0.217} & 848/16\\
Fixed, $\beta=1.21$ & 88.3 & 0.226 & 848/16\\
Fixed, $\beta=2$ & 91.7 & 0.225 & 848/16\\
Fixed, $\beta=5$ & 89.2 & 0.227 & 848/16\\
\midrule
\multicolumn{4}{@{}l}{\emph{PIE-Bench: preview length and target}}\\
Configuration & EditRate$\uparrow$ & LPIPS$\downarrow$ & U-Net/judge\\
$L=2,\ \tau=15$ & 18.5 & 97.9 & 288/16\\
$L=8,\ \tau=15$ & \textbf{19.3} & \textbf{97.8} & 432/16\\
$L=2,\ \tau=18$ & 18.9 & 97.9 & 288/16\\
\bottomrule
\end{tabular}
\caption{\textbf{Sensitivity of image editing to preview and steering settings.} Four particles, $240$ valid SEAM-Bench sources; PIE-Bench uses $s=0.3$. EditRate is a percentage and LPIPS is multiplied by $10^3$. Bold marks the best metric within each block.}
\label{tab:supp-preview-edit}
\end{table}

\vspace{5pt}
\noindent
\textbf{Steering improves edit success at a fixed source-noise level.}
\label{sec:supp-sdedit}
\label{sec:st-sdedit}
On PIE-Bench, increasing the SDEdit noise strength of Single improves edit success but progressively reduces source preservation (\cref{tab:supp-sdedit}). At the default noise strength, \methodname{} improves edit success over Single with comparable preservation, illustrating a different success--preservation tradeoff than increasing source noise alone.

\begin{table}[!ht]
\centering
\scriptsize
\setlength{\tabcolsep}{2pt}
\begin{tabular}{@{}lccccc@{}}
\toprule
Method / $s$ & EditRate$\uparrow$ & Struct.$\downarrow$ & PSNR$\uparrow$ & LPIPS$\downarrow$ & U-Net/judge\\
\midrule
\multicolumn{6}{@{}l}{\emph{Main results ($s=0.3$)}}\\
Single & 13.4 & \textbf{18.38} & 24.05 & 98.0 & $60/1$\\
\methodname{} & \textbf{19.3} & 18.43 & \textbf{24.08} & \textbf{97.8} & $432/16$\\
\midrule
\multicolumn{6}{@{}l}{\emph{Single: independent sweep}}\\
$0.3$ & 13.6 & \textbf{18.42} & \textbf{24.07} & \textbf{97.9} & $60/1$\\
$0.4$ & 20.8 & 25.42 & 22.62 & 125.1 & $80/1$\\
$0.5$ & 28.8 & 34.89 & 21.13 & 157.1 & $100/1$\\
$0.6$ & 32.7 & 47.88 & 19.56 & 190.6 & $120/1$\\
$0.7$ & 44.2 & 64.68 & 17.88 & 228.5 & $140/1$\\
$0.8$ & 47.9 & 86.81 & 16.22 & 265.5 & $160/1$\\
$0.9$ & \textbf{51.9} & 116.97 & 14.48 & 307.4 & $180/1$\\
\bottomrule
\end{tabular}
\caption{\textbf{SDEdit noise strength on PIE-Bench.} SD v1.4. Main-experiment results are shown alongside an independent Single noise-strength sweep. EditRate is a percentage; Struct. (Structure) and LPIPS are multiplied by $10^3$. Bold marks the best score within each block.}
\label{tab:supp-sdedit}
\end{table}

\vspace{5pt}
\noindent
\textbf{Steering improves counter-stereotypical editing over four-particle baselines while maintaining low asymmetry.}
\label{sec:supp-bias}
We report success in both editing directions using the source-validity protocol in \cref{sec:supp-seam-metrics}. \methodname{} improves against-stereotype edit success over the four-particle baselines and achieves the smallest absolute directional asymmetry on both backbones, including a tie on SD v1.4 (\cref{tab:supp-bias}). This supports more balanced editing across the tested associations rather than uniformly higher success in both directions.

\begin{table}[!ht]
\centering
\scriptsize
\setlength{\tabcolsep}{3pt}
\begin{tabular}{@{}lccc@{}}
\toprule
& \multicolumn{2}{c}{Directional rates} & Shared contexts\\
\cmidrule(lr){2-3}\cmidrule(lr){4-4}
Method & Toward$\uparrow$ & Against$\uparrow$ & Signed asym.\\
\midrule
\multicolumn{4}{@{}l}{\emph{SD v1.4} ($66$ toward, $80$ against, $57$ shared)}\\
Single & 89.4 & 78.8 & $+5.3$\\
Best-of-4 & 95.5 & 82.5 & $\mathbf{+1.8}$\\
Best-of-8 & \textbf{97.0} & \textbf{85.0} & $+7.0$\\
FK steering & 89.4 & 77.5 & $+5.3$\\
\methodname{} & 90.9 & 83.8 & $\mathbf{-1.8}$\\
\midrule
\multicolumn{4}{@{}l}{\emph{SDXL} ($74$ toward, $90$ against, $69$ shared)}\\
Single & 89.2 & 76.7 & $+4.3$\\
Best-of-4 & \textbf{100.0} & 83.3 & $+5.8$\\
Best-of-8 & 98.6 & 83.3 & $+4.3$\\
FK steering & 98.6 & 81.1 & $+5.8$\\
\methodname{} & 95.9 & \textbf{90.0} & $\mathbf{-1.4}$\\
\bottomrule
\end{tabular}
\caption{\textbf{Counter-stereotypical editing.} Directional EditRates use their own valid sources; signed asymmetry uses shared-valid contexts. Rates are percentages, differences are percentage points. Bold marks highest directional rates and smallest absolute signed asymmetry within each model.}
\label{tab:supp-bias}
\end{table}
\setcounter{dbltopnumber}{5}
\renewcommand{\dbltopfraction}{0.95}
\renewcommand{\dblfloatpagefraction}{0.75}
\section{Additional experiments}
\label{sec:supp-additional-experiments}

We provide strength settings and additional metrics for the editing baselines in the main paper, compare inference-time sampling with fine-tuned generators on compositional benchmarks, and report wall-clock inference times.

\subsection{Guidance and attention-control baselines}
\label{sec:supp-additional-baselines}

We report the editing configurations for Prompt-to-Prompt, SEGA, Universal Guidance, and FreeDoM, using SD v1.4 and the evaluation protocols in \cref{sec:supp-seam-settings,sec:supp-seam-metrics,sec:supp-edit-eval}. Each method returns one edited image per input. For PIE-Bench, these methods start from the source-conditioned SDEdit initialization used by Single and run the final $30$ denoising steps, with their schedules scaled to the shorter trajectory.

\vspace{5pt}
\noindent
\textbf{Strength settings.}
\label{sec:supp-edit-baseline-settings}
We evaluate the strength grids in \cref{tab:supp-edit-baseline-settings} and report the setting with the highest EditRate, breaking ties by lower preservation error $E$ on \dsEdit{} or lower LPIPS on PIE-Bench. These settings are selected on the evaluation sets, whereas the sampling baselines retain fixed configurations. The $100$ scoring calls for Universal Guidance and FreeDoM are CLIP energy evaluations, not reward-judge calls.

\begin{table}[!ht]
\centering
\scriptsize
\setlength{\tabcolsep}{2pt}
\begin{tabular}{@{}llcc@{}}
\toprule
Method & Strength grid & SEAM & PIE\\
\midrule
Prompt-to-Prompt & $\{0.2,0.4,0.6,0.8,1\}$ & 1 & 1\\
SEGA & $\{3,5,7,10,14,20\}$ & 20 & 20\\
Universal Guidance & $\{1,2.5,5,10,15,20\}$ & 15 & 15\\
FreeDoM & $\{0.2,0.5,1,2,4,8\}$ & 2 & 8\\
\bottomrule
\end{tabular}
\caption{\textbf{Editing baseline strengths (SD v1.4).} Tested grids and selected strengths on SEAM-Bench (SEAM) and PIE-Bench (PIE).}
\label{tab:supp-edit-baseline-settings}
\vspace{-10pt}
\end{table}

\vspace{5pt}
\noindent
\textbf{\methodname{} achieves the highest edit success on \dsEdit{}.}
\methodname{} also achieves lower overall preservation error than the other sampling methods in \cref{tab:edit-baselines}. Prompt-to-Prompt better preserves source content but completes fewer of the requested edits, illustrating the importance of evaluating success and preservation jointly.

\begin{table*}[t]
\centering
\footnotesize
\setlength{\tabcolsep}{5pt}
\begin{tabular}{@{}l c ccccc@{}}
\toprule
Method & \shortstack{Particles\\$n$} & EditRate$\uparrow$ & $E\downarrow$ & $E_{\mathrm{verified}}\downarrow$ & \shortstack{Context\\lost}$\downarrow$ & \shortstack{U-Net /\\judge calls}\\
\midrule
\multicolumn{7}{@{}l}{\emph{Sampling and steering}}\\
Single & 1 & 80.4 & 0.225 & 0.223 & 6.2 & 200\,/\,1\\
Best-of-4 & 4 & 88.8 & 0.219 & 0.220 & 4.2 & 800\,/\,4\\
Best-of-8 & 8 & 91.2 & 0.219 & 0.221 & \textbf{2.1} & 1600\,/\,8\\
FK steering~\cite{singhal2025fk} & 4 & 87.5 & 0.224 & 0.218 & 4.6 & 800\,/\,24\\
\methodname{} & 4 & \textbf{92.5} & 0.217 & 0.218 & 2.9 & 992\,/\,16\\
\midrule
\multicolumn{7}{@{}l}{\emph{Guidance and attention methods}}\\
Prompt-to-Prompt & 1 & 73.8 & \textbf{0.165} & \textbf{0.172} & 4.2 & 400\,/\,0\\
SEGA & 1 & 80.4 & 0.225 & 0.223 & 6.2 & 400\,/\,0\\
Universal Guidance & 1 & 25.8 & 0.340 & 0.330 & 25.0 & 400\,/\,100$^\dagger$\\
FreeDoM & 1 & 37.1 & 0.371 & 0.384 & 16.2 & 520\,/\,100$^\dagger$\\
\bottomrule
\end{tabular}
\caption{\textbf{Editing comparisons on \dsEdit{}.} SD v1.4. EditRate and Context lost are percentages; $\dagger$ marks CLIP energy evaluations. Bold indicates the best value in each metric.}
\label{tab:edit-baselines}
\end{table*}

\vspace{5pt}
\noindent
\textbf{\methodname{} improves edit success over Single while maintaining comparable source preservation.}
On PIE-Bench, the preservation metrics remain similar to those of Single despite improved edit success. SEGA and FreeDoM achieve higher edit success but introduce larger structural and perceptual changes (\cref{tab:edit-baselines-pie}).

\begin{table}[!ht]
\centering
\scriptsize
\setlength{\tabcolsep}{1.7pt}
\begin{tabular}{@{}l c ccccc@{}}
\toprule
Method & $n$ & \shortstack{EditRate$\uparrow$} & $\mathrm{CLIP}_{\mathrm{edit}}\uparrow$ & Struct.$\downarrow$ & PSNR$\uparrow$ & LPIPS$\downarrow$\\
\midrule
\multicolumn{7}{@{}l}{\emph{Sampling and steering}}\\
Single & 1 & 13.4 & 21.35 & 18.38 & 24.05 & 98.0\\
Best-of-4 & 4 & 18.5 & 21.60 & 18.59 & 24.02 & 98.4\\
Best-of-8 & 8 & 21.0 & 21.69 & 18.49 & 24.01 & 97.9\\
FK steering~\cite{singhal2025fk} & 4 & 20.6 & 21.69 & 18.53 & 24.04 & 98.1\\
\methodname{} & 4 & 19.3 & 21.65 & 18.43 & 24.08 & 97.8\\
\midrule
\multicolumn{7}{@{}l}{\emph{Guidance and attention methods}}\\
Prompt-to-Prompt & 1 & 12.3 & 21.33 & 18.34 & 24.07 & 97.7\\
SEGA & 1 & 27.0 & 21.69 & 20.37 & 23.61 & 103.6\\
Universal Guidance & 1 & 5.6 & 21.47 & \textbf{13.90} & \textbf{25.19} & \textbf{82.0}\\
FreeDoM & 1 & \textbf{29.8} & \textbf{21.85} & 135.41 & 15.40 & 335.4\\
\bottomrule
\end{tabular}
\caption{\textbf{Editing comparisons on PIE-Bench.} SD v1.4. EditRate is a percentage; Struct. (Structure) and LPIPS are multiplied by $10^3$. Bold indicates the best value in each metric.}
\label{tab:edit-baselines-pie}
\vspace{-10pt}
\end{table}

\subsection{Finetuning}
\label{sec:supp-finetune}

We compare fine-tuning and inference-time sampling on the two compositional generation benchmarks, \dsGen{} and T2I-CompBench++. We use two publicly released checkpoints: \href{https://huggingface.co/kvablack/ddpo-alignment}{DDPO}~\cite{black2024ddpo} for SD v1.4, trained with reinforcement learning on animal-activity prompts, and \href{https://huggingface.co/mhdang/dpo-sdxl-text2image-v1}{Diffusion-DPO}~\cite{wallace2024diffusiondpo} for SDXL, trained on preference pairs. We replace only the U-Net and retain the base model's remaining components and sampling settings (\cref{sec:supp-sampling}).

\vspace{5pt}
\noindent
\textbf{Evaluation.}
For each prompt, we generate four images from the base and fine-tuned models using the same initial noise samples as Best-of-4. To estimate performance without selection, we average the independent benchmark scores of the four images. To evaluate Best-of-4, we select the image with the highest ImageReward score and report its benchmark score. We compare these configurations with FK steering and \methodname{} applied to the base model. For unselected configurations, the four-sample mean estimates single-image performance, and the reported cost is per image. For selection-based methods, costs include all candidates used to return one image.

\vspace{5pt}
\noindent
\textbf{\methodname{} outperforms Best-of-4 sampling from the tested fine-tuned generators.}
Reward-based selection improves alignment for both fine-tuned generators, yet \methodname{} achieves higher scores on both compositional benchmarks across both backbones (\cref{tab:finetune}). These results demonstrate the benefits of adaptive inference-time steering without modifying the pretrained model parameters.

\begin{table}[!ht]
\centering
\scriptsize
\setlength{\tabcolsep}{2.7pt}
\begin{tabular}{@{}lccc@{}}
\toprule
Method & \dsGen$\uparrow$ & T2I$\uparrow$ & U-Net/judge\\
\midrule
\multicolumn{4}{@{}l}{\emph{SD v1.4}}\\
Base model (mean) & 44.0 & 35.2 & 200/1\\
DDPO (mean)~\cite{black2024ddpo} & 42.8 & 38.1 & 200/1\\
DDPO + Best-of-4 & 52.9 & 47.3 & 800/4\\
Base model + Best-of-4 & 53.9 & 46.1 & 800/4\\
FK steering~\cite{singhal2025fk} & 53.0 & 44.7 & 800/24\\
\methodname{} & \textbf{56.3} & \textbf{50.7} & 928/12\\
\midrule
\multicolumn{4}{@{}l}{\emph{SDXL}}\\
Base model (mean) & 55.6 & 47.6 & 200/1\\
Diffusion-DPO (mean)~\cite{wallace2024diffusiondpo} & 60.1 & 52.8 & 200/1\\
Diffusion-DPO + Best-of-4 & 66.2 & 57.8 & 800/4\\
Base model + Best-of-4 & 63.0 & 54.2 & 800/4\\
FK steering~\cite{singhal2025fk} & 61.4 & 54.4 & 800/24\\
\methodname{} & \textbf{69.4} & \textbf{58.4} & 928/12\\
\bottomrule
\end{tabular}
\caption{\textbf{Fine-tuned generators and inference-time sampling.} Independent evaluation (0--100). Four-sample means estimate unselected performance; Best-of-4 uses ImageReward. Costs for unselected models are per generated image; selection costs are per returned image. T2I denotes T2I-CompBench++; bold marks the best score for each model.}
\label{tab:finetune}
\vspace{-10pt}
\end{table}

\subsection{Inference time}
\label{sec:supp-timing}

We measure wall-clock time to complement U-Net and reward-judge counts, which do not fully capture batching and model-dependent scoring costs.

\vspace{5pt}
\noindent
\textbf{Setup.}
We measure inference time for Single, Best-of-4, FK steering, and \methodname{} using SD v1.4 in half precision on one NVIDIA A100 80GB GPU. We use the main $100$-step configurations, ImageReward for generation and Qwen2.5-VL-7B for \dsEdit{} editing. Each method is timed on $24$ \dsGen{} prompts and $24$ \dsEdit{} edits, after model loading and two untimed warm-up runs. We synchronize the GPU around each query and judge call, batch the particles within each four-particle method, and exclude disk writes. Judge time measures reward scoring; total time also includes denoising, lookaheads, and other sampling operations.

\vspace{5pt}
\noindent
\textbf{\methodname{} reduces reward-scoring time with similar total latency to FK steering.}
In both generation and editing, the lower scoring time helps offset the additional computation required by lookahead previews (\cref{tab:supp-timing}).

\begin{table}[t]
\centering
\scriptsize
\setlength{\tabcolsep}{3pt}
\begin{tabular}{@{}lcccc@{}}
\toprule
& \multicolumn{2}{c}{\shortstack{Generation\\(ImageReward)}} & \multicolumn{2}{c}{\shortstack{\dsEdit{}\\(Qwen2.5-VL)}}\\
\cmidrule(lr){2-3}\cmidrule(lr){4-5}
Method & Total (s) & Judge (s) & Total (s) & Judge (s)\\
\midrule
Single & $\mathbf{4.7\pm1.0}$ & -- & $\mathbf{4.3\pm1.0}$ & --\\
Best-of-4 & $7.7\pm0.1$ & \textbf{0.2} & $8.3\pm0.1$ & \textbf{0.7}\\
FK steering & $8.9\pm0.2$ & 0.9 & $12.3\pm0.4$ & 4.2\\
\methodname{} & $9.2\pm0.1$ & 0.5 & $12.3\pm0.2$ & 2.9\\
\bottomrule
\end{tabular}
\caption{\textbf{Inference time on one A100 80GB.} Mean seconds over $24$ queries per task; total time includes standard deviation. Bold marks the lowest mean time; judge time is compared only where measured. Single is timed without reward scoring; its reported judge time is unavailable.}
\label{tab:supp-timing}
\vspace{-10pt}
\end{table}
\section{Limitations}
\label{sec:supp-limitations}
As with other reward-guided methods, \methodname{} depends on the quality of its reward judge. We evaluate outputs using independent models and benchmark metrics, but these evaluations cannot fully capture all aspects of image quality or prompt alignment. Lookahead previews also introduce additional computation, with the tradeoff between denoising and judge evaluations depending on the inference setting. Finally, our counter-stereotypical editing experiments assess specific attribute changes and do not establish broader fairness properties of the model or its outputs.
\section{Additional Related Work}
\label{sec:supp-add-related-work}

The main text focuses on inference-time methods for improving text-image alignment. Here, we discuss related work on image editing and generative debiasing that provides additional context for our editing experiments.

Image editing aims to modify specified attributes while preserving unrelated scene content. Existing approaches use source-image noising~\cite{meng2022sdedit}, attention control~\cite{hertz2022prompt}, and feature injection or inversion~\cite{tumanyan2023pnp,mokady2023nulltext,ju2024pnp}. PIE-Bench evaluates editing methods in terms of both edit success and source preservation~\cite{ju2024pnp}. Related work on generative debiasing seeks to reduce unwanted associations through guidance~\cite{friedrich2023fairdiffusion}, learned embeddings~\cite{zhang2023itigen}, or finetuning~\cite{shen2024finetuning}, while BendVLM studies training-free interventions on vision-language representations~\cite{gerych2024bendvlm}. Together, these works provide context for our evaluation of whether inference-time particle steering can achieve targeted edits, including counter-stereotypical changes, while preserving unrelated image content.

\end{document}